%% file: main.tex
\documentclass[letterpaper]{article} % DO NOT CHANGE THIS
\usepackage{aaai2026}  % DO NOT CHANGE THIS
\usepackage{times}  % DO NOT CHANGE THIS
\usepackage{helvet}  % DO NOT CHANGE THIS
\usepackage{courier}  % DO NOT CHANGE THIS
\usepackage[hyphens]{url}  % DO NOT CHANGE THIS
\usepackage{graphicx} % DO NOT CHANGE THIS
\usepackage{natbib}  % DO NOT CHANGE THIS AND DO NOT ADD ANY OPTIONS TO IT
\usepackage{caption} % DO NOT CHANGE THIS AND DO NOT ADD ANY OPTIONS TO IT
\usepackage{algorithm}
\usepackage{algorithmic}
\usepackage{amsmath}
\usepackage{tabularx}
\usepackage{newfloat}
\usepackage{listings}
\DeclareCaptionStyle{ruled}{labelfont=normalfont,labelsep=colon,strut=off} % DO NOT CHANGE THIS
\floatstyle{ruled}
\newfloat{listing}{tb}{lst}{}
\floatname{listing}{Listing}
\usepackage{booktabs}
\usepackage{makecell}
\usepackage{listings}
\usepackage{multirow}

\title{Prune4Web: DOM Tree Pruning Programming for Web Agent}
\author{
    Jiayuan Zhang\equalcontrib,
    Kaiquan Chen\equalcontrib,
    Zhihao Lu,
    Enshen Zhou,
    Qian Yu,
    Jing Zhang\thanks{Corresponding author: Jing Zhang}
}
\affiliations{
    School of Software \& QRI, Beihang University, Beijing, China\\
    \{zhangjiayuan42, zhang\_jing\}@buaa.edu.cn
}

\begin{document}

\maketitle

\begin{abstract}
Web automation uses intelligent agents to perform high-level tasks by mimicking human interactions with webpages. Despite recent advances in LLM-based web agents, efficiently navigating complex, real-world webpages remains challenging due to massive DOM structures (10,000$\sim$100,000 tokens). Current approaches either truncate DOMs—losing vital information—or use inefficient heuristics and separate ranking models, failing to balance precision and scalability. We introduce \textbf{Prune4Web}, a novel paradigm that transforms DOM processing from LLM-based filtering to programmatic pruning. Our key innovation is DOM Tree Pruning Programming, where an LLM generates executable Python scoring programs to dynamically filter DOM elements based on semantic clues from decomposed sub-tasks. This approach eliminates the need for LLMs to process full DOMs, instead delegating traversal and scoring to lightweight, interpretable programs. The result is a \textbf{25$\sim $50 times reduction} in candidate elements for grounding, enabling precise action localization without attention dilution. Additionally, we propose a data annotation method and a two-turn dialogue training strategy that jointly optimizes Planner, Programmatic Filter, and Grounder in a unified framework. Experiments demonstrate state-of-the-art performance. On our low-level task grounding task, our approach dramatically increases grounding accuracy from \textbf{46.8\% to 88.28\%}, highlighting its effectiveness.
\end{abstract}


\section{Introduction}
\label{sec:introduction}

Web automation enables the completion of high-level tasks, such as booking flights or shopping online, through intelligent agents that mimic human interaction on webpages. These agents achieve this by interpreting high-level tasks, breaking them down into low-level sub-tasks, and seamlessly interacting with web elements. Recently, large language models (LLMs) have demonstrated impressive capabilities in autonomous web navigation through their strong reasoning and decision-making abilities~\cite{yao2022webshop, deng2023mind2web}. Current web agents approaches fall into three main categories: 1) Textual HTML/DOM-based~\cite{yao2022webshop, song2024beyond}, 2) Visual Screenshot-based~\cite{lin2024showui, cheng2024seeclick}, and 3) Multi-modal-based~\cite{he2024webvoyager, zheng2024gpt}. Visual screenshots provide an intuitive, human-like understanding of webpage state, making them effective for reasoning about low-level sub-tasks. However, they contain limited semantic information, especially for special icons, and are sensitive to variations in resolution and overlapping elements. In contrast, HTML/DOMs offer precise and stable semantic and structural information that enables accurate element selection with minimal ambiguity.

\begin{figure}[t]
    % \centering
    \captionsetup{type=figure} \includegraphics[width=\linewidth]{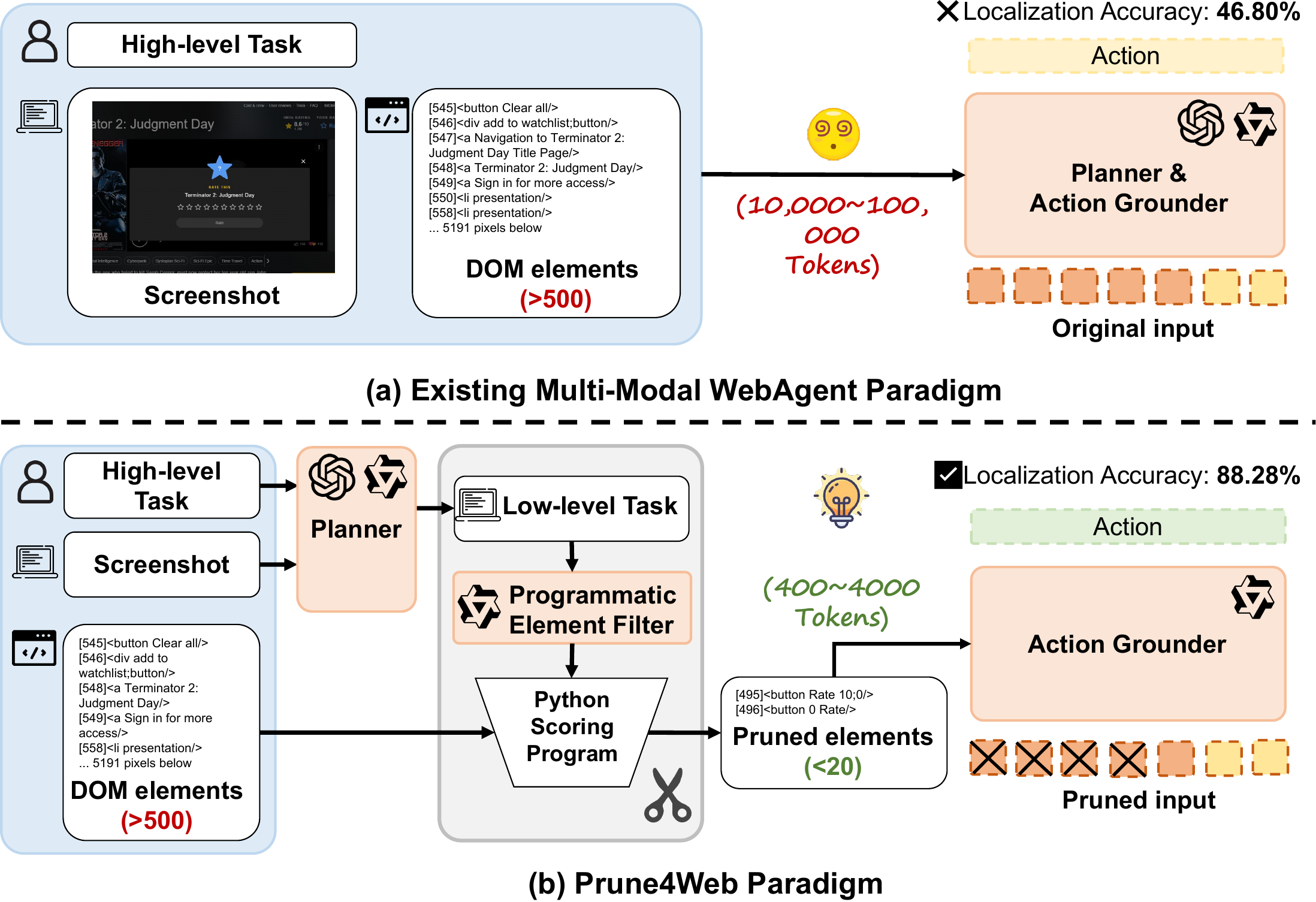}
    \captionof{figure}{Comparison between existing multi-modal web agents and our Prune4Web paradim. Compared to existing multi-modal web agent paradigms, we propose a programmatic pruning strategy that efficiently removes redundant DOM elements. Our Prune4Web approach relaxes the token limits of LLMs and increases accuracy on low-level sub-task grounding from 46.80\% to 88.28\%.}
    \label{fig:teaser}
\end{figure}

In this paper, we leverage the complementary advantages of text and visual multi-modal information and design a multi-stage framework: A \textbf{planner} model takes the high-level task (e.g., ``Book a flight to New York'') and a \textbf{screenshot}, then decomposes it into a low-level sub-task (e.g., ``Find the destination field and Type NYC''). Based on the sub-task, an \textbf{action grounder} model processes the \textbf{DOMs} to precisely localize and execute the required operations (e.g., selecting \textless input id=``destination''\textgreater to type ``NYC''). However, modern webpage DOMs typically contain 10,000–100,000 tokens—far exceeding the context capacity of most LLMs. This results in token truncation and attention dilution, leading to critical information loss and significant processing delays~\cite{gou2024navigating, deng2023mind2web}. Existing HTML pruning methods fall short, either relying on overly simplistic heuristic filtering~\cite{he2024webvoyager, pan2024webcanvas} or requiring separate language models for element-ranking~\cite{deng2023mind2web}. Neither approach effectively addresses the core issue. The fundamental challenge remains: \emph{how to efficiently and accurately navigate task-relevant elements from complete DOM structures.}

To this end, we propose a \textbf{Prune4Web} pipeline through a novel paradigm: DOM Tree Pruning Programming. We observe that the low-level sub-tasks (e.g.,``Find the destination field'') output by the planner contain extensive semantic clues about potentially relevant DOM elements. This insight motivates us to shift the LLM's role from directly locating elements in lengthy DOMs to generating a locator program based solely on the low-level sub-tasks, thereby avoiding the need to feed long DOM sources into the LLMs~\cite{jiang2024survey, zhang2023planning}. Specifically, we implement this concept through our \textbf{Programmatic Element Filter} model. This filter receives a specific low-level sub-task from the upstream Planner and prompts the LLM to generate a concise, task-specific Python \textit{scoring program}. We design a heuristic-based scoring program template, requiring the LLM to generate only key parameters for better controllability and flexibility. The generated program runs independently outside the LLM, efficiently traversing the complete DOM tree to score and rank all elements. This approach reduces candidate elements by \textbf{25$\sim $50 times}, enabling precise action localization without attention dilution. A downstream LLM-based Action Grounder then selects the final element from this refined shortlist, completing the grounding task.

To train the models within Prune4Web, we create an automated data synthesis pipeline that annotates structured intermediate outputs from raw data with minimal human intervention. These include low-level sub-tasks for the Planner and key parameters for the Programmatic Element Filter. For optimization, we develop a novel two-turn dialogue training strategy that jointly trains the Planner, Filter, and Grounder as a unified model. We initially use Supervised Fine-Tuning (SFT) with our annotated data to train a base model~\cite{zheng2024llamafactory}. Subsequently, we apply Reinforcement Fine-Tuning (RFT) to enhance the Planner's long-term planning capabilities while integrating the programmatic filtering process into this optimization framework. Extensive experiments on benchmark datasets~\cite{deng2023mind2web, pan2024webcanvas} demonstrate the effectiveness of the proposed Prune4Web. Notably, on our low-level sub-task grounding benchmark, our approach greatly boosts grounding accuracy from \textbf{46.8\%} to \textbf{88.28\%}, showing its core advantage.
%
% In summary, our main contributions include:
%
Our contributions are summarized as follows:
\begin{itemize}
    \item We design a multimodal web agent framework that seamlessly combines the intuitive reasoning of visual inputs with the semantic precision of HTML/DOM.
    \item We introduce Prune4Web with a Programmatic Element Filter that generates task-specific Python scoring programs to efficiently filter and rank elements to address the DOM scalability bottleneck.
    \item We present a data annotation method and a two-turn dialogue training strategy that jointly optimize the planner, filter, and grounder. We use SFT and RFT to enhance planning and programmatic filtering. Strong empirical evidence validates our method on standard benchmarks.
\end{itemize}

\section{Related Work}
\label{sec:related_work}

\noindent\textbf{Multimodal-based Web Agents.}
To achieve higher precision in web interaction, directly processing HTML source code has recently become a significant research direction for LLM-based agents~\cite{lai2024autowebglm, song2024beyond}, leading to notable advancements. Researchers have leveraged the rich semantic and structural information within the DOM by developing multimodal fusion techniques~\cite{zheng2024gpt, furuta2023multimodal} or more powerful end-to-end models~\cite{lin2024showui, cheng2024seeclick, hong2024cogagent, xu2024aguvis} for precise element localization and operation. However, these efforts toward precision inevitably face the challenge of information overload~\cite{gou2024navigating, deng2023mind2web, xue2025illusion}. Modern webpage HTML sources typically contain vast amounts of irrelevant information. Feeding this directly into an LLM wastes computational resources and dilutes the model's focus across lengthy context~\cite{gur2023real}. Balancing HTML's precision with efficient information processing remains a critical, unsolved challenge.

\noindent\textbf{DOM Tree Pruning Strategies.}
DOM tree pruning is a key technique for addressing information overload challenge. Existing methods fall into two categories. The first is rule-based filtering, which relies on fixed heuristics like converting the DOM to a simplified accessibility tree~\cite{he2024webvoyager,  zhou2023webarena}. The second is LLM-based ranking, where the model is prompted to score and select from a large number of element candidates~\cite{deng2023mind2web, lu2024weblinx,kerboua2025lineretriever}. Rule-based approaches are too rigid and generalize poorly. LLM-based ranking fails to reduce the burden of processing long contexts. In contrast, our work introduces DOM Tree Pruning Programming, a new paradigm that addresses both limitations by having the LLM generate a lightweight locator program~\cite{qiao2024autoact, jiang2024self}.

\noindent\textbf{Programmatic Thinking for Agents.}
Our method is rooted in programmatic thinking—a paradigm that enhances LLM abilities by prompting them to generate intermediate code or plans to solve complex problems. This approach has shown effective for general reasoning and planning~\cite{jiang2024survey, zhang2023planning, wei2022chain, gupta2023visual, zhou2025code, qin2024mp5}. In the agent field, programmatic thinking typically generates high-level action sequences that control agent behavior on the web~\cite{ma2023laser}, mobile devices~\cite{wen2024autodroid, zhang2023appagent}, or general computer operations~\cite{zhang2024ufo, tan2024cradle, wu2024copilot, xie2025osworld}. Our work innovatively applies this paradigm to the lower-level problem of DOM filtering by generating an executable scoring function that actively reshapes the model's input.

\noindent\textbf{Reinforcement Fine-Tuning for Agents.}
To enable agents to learn complex policies beyond static datasets, Reinforcement Fine-Tuning (RFT) is increasingly used to optimize LLM agents for sequential decision-making in dynamic environments~\cite{lu2025uir1, qi2024webrl, bai2025digirl, guo2025deepseek}. RFT allows agents to learn from outcomes via a reward mechanism, enabling them to master complex strategies. In Prune4Web, we not only employ RFT to optimize the Planner's capabilities but also innovatively use the success or failure of our DTPP process to provide rich intermediate reward signals~\cite{zhou2025roborefer}, facilitating more efficient policy learning.

\begin{figure*}
    \centering
    \includegraphics[width=1\linewidth]{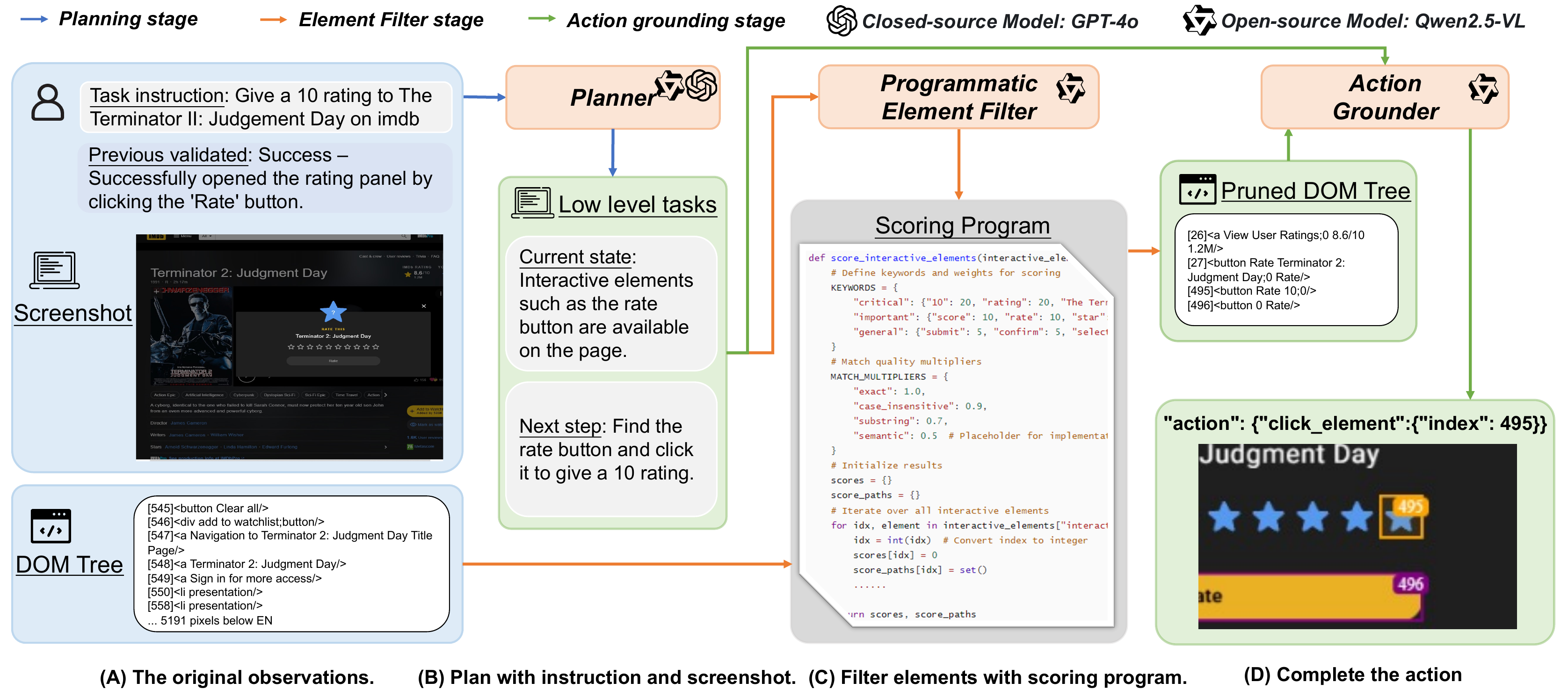}
    \caption{The Prune4Web framework pipeline.
(A) The input observations include high-level task, history, screenshot, and DOM tree.
(B) The Planner generates low-level subtasks based on user instruction, history, and screenshot.
(C) Guided by the plan, the Programmatic Element Filter produces a scoring program that is applied to the DOM tree elements to yield a pruned DOM tree.
(D) This pruned DOM forms the refined input for the Action Grounder, which then selects an executable action.}
    \label{fig:pipeline}
\end{figure*}

\section{Method}
\label{sec:method}

\subsection{Prune4Web Framework and Workflow}
\label{subsec:framework}

We introduce Prune4Web, a multi-stage framework for complex web automation tasks. The complete workflow is shown in Figure~\ref{fig:pipeline}. The framework consists of three stages: task planning, element filtering, and action grounding.
% The core idea is twofold: first, to leverage the precise and rich information from HTML source code to overcome the limitations of purely visual analysis, and second, to systematically resolve the information overload introduced by HTML itself through a programmatic filtering stage.

\noindent\textbf{Planning Stage.}
The workflow begins with the Planner model, which decomposes a high-level task $T$ into low-level sub-tasks $S_t$ based on the current webpage screenshot $Sc_t$ and operational history $H_t$. This process is formally expressed as: $S_t = \text{Planner}(T, Sc_t, H_t)$. For example, given the high-level task ``Book a flight to New York,'' the Planner might generate low-level sub-tasks like ``Find the destination field'' and ``Type NYC'', as well as current states. The Planner intentionally does not access the HTML source code, keeping its focus on high-level strategic decomposition.

\noindent\textbf{Filtering Stage.}
If a low-level sub-task $S_t$ requires interaction with a specific element, the workflow proceeds to the filtering stage managed by the Programmatic Element Filter model. This model implements our core method, DOM Tree Pruning Programming, to generate a refined list of candidate elements $C_t$ from the complete HTML source code: $C_t = \text{ProgrammaticElementFilter}(S_t, \text{HTML}_t)$. The resulting list $C_t$ then serves as the sole input for the subsequent Action Grounder.

\noindent\textbf{Action Grounding Stage.}
The Action Grounder completes the workflow by generating the final executable action $A_t$. It takes two inputs: the low-level sub-task $S_t$ from the Planner and the pruned candidate list $C_t$ from the Programmatic Element Filter. This is formally expressed as $A_t = \text{ActionGrounder}(S_t, [C_t])$, where the brackets indicate that $[C_t]$ is conditional. This is because $[C_t]$ is only required for element-specific actions (e.g., `click'), whereas abstract actions (e.g., `task complete') are grounded directly from $S_t$.

\noindent In summary, the Prune4Web framework offers a dual advantage. It uses the structured, precise information from the DOM to avoid the pitfalls of visual-based localization in complex scenarios. At the same time, its innovative filtering stage distills verbose HTML into a concise list of candidates. This effectively mitigates information overload and significantly reduces the difficulty and error rate of the grounding task for the Action Grounder.

\subsection{DOM Tree Pruning Programming}
\label{subsec:core_method}

DOM Tree Pruning Programming is the technical core of Prune4Web. It offloads the heavy task of element filtering from the LLM itself to a lightweight, dynamically generated program.

\noindent\textbf{Step 1: Initial Rule-based Filtering.}
The process begins with a rule-based preliminary filtering of the raw $\text{HTML}_t$. The core principle is to retain elements with clear interactive features based on their tags (e.g., \textless a\textgreater, \textless button\textgreater, \textless input\textgreater) or `role' attributes (e.g., `checkbox'). For non-interactive elements, we extract key textual information (from `text', `aria-label', etc.) and attach it to the nearest interactive element as supplementary context. This step yields a pre-processed DOM tree containing only context-enriched interactive elements, serving as a more structured and less noisy initial candidate set.

\noindent\textbf{Step 2: Scoring Function Generation.}
The core task of the Programmatic Element Filter is to generate a Python scoring function $f_{score\_t}$ for the current step. We design a heuristic-based Scoring Function Template, where the LLM only needs to generate key parameters for this template. This approach significantly improves the stability and controllability of the generated code while maintaining flexibility. Algorithm~\ref{alg:context_aware_scoring} shows the pseudo-code of the template. The template mimics human intuition when searching for elements using keywords. It assumes that a target element contains identifiable textual features within the HTML. The template performs tiered, weighted matching across different attributes: Tier 1 includes visible `text'; Tier 2 includes non-visible but high-semantic attributes like `aria-label' and `placeholder'; and Tier 3 includes other attributes like `class' or `id' that may contain semantic cues. The template also integrates multiple matching types (e.g., exact, substring, fuzzy) and assigns weights based on match quality. With this design, the Programmatic Element Filter simply generates a set of keywords and their corresponding base weights based on the low-level sub-task $S_t$, enabling multi-faceted relevance scoring for each element.

% \noindent\textbf{Step 3: Pruning Execution and Output Formatting.}
% The generated scoring function $f_{score\_t}$ is immediately executed to compute a score $s$ for each element $e$ in the pre-processed DOM tree. Subsequently, the system selects the Top-N highest-scoring elements. The value of N is determined dynamically, as the number of elements and the variance of scores differ drastically across webpages and steps. Dynamic selection based on the score distribution allows for a better trade-off between information volume and completeness. The value of N is calculated using the formula $N = \min(N_{max}, \max(N_{min}, \lceil \mu_s + \alpha \cdot \sigma_s \rceil))$, where the hyperparameters were determined through ablation studies.

\noindent\textbf{Step 3: Pruning Execution and Output Formatting.}
The generated scoring function $f_{score\_t}$ is immediately executed to compute a score $s$ for each element $e$ in the pre-processed DOM tree. The system then selects the Top-N highest-scoring elements, where N defaults to 20. The impact of varying N from 1 to 20 on pruning efficiency is visualized and analyzed in the Experiments section.

\noindent\textbf{Discussion.} The primary advantage of DOM Tree Pruning Programming lies in its combination of flexibility and structure. The LLM provides high-level, context-aware intelligence by generating keywords and weights, while the hard-coded function template ensures robust, efficient, and interpretable scoring and execution. By generating a lightweight function instead of directly processing lengthy raw HTML, this paradigm avoids attention dilution from long contexts and significantly reduces inference latency.

\subsection{Data Synthesis}
\label{subsec:data_synthesis}

To effectively train our multi-stage architecture and rigorously evaluate DOM Tree Pruning Programming, we reconstructed and re-annotated the public Multimodal-Mind2Web (MM2W) dataset~\cite{deng2023mind2web}. The original MM2W dataset contains only high-level tasks, per-step source code, and final target elements and actions, lacking the intermediate reasoning steps our framework requires. To address this, we used GPT-4o~\cite{hurst2024gpt} as an annotation tool to add rich intermediate labels for each step. These labels include: (1) low-level sub-tasks for the Planner; (2) keywords and their weights for the Programmatic Element Filter; and (3) pruned DOM trees and thought processes for the Action Grounder. After annotation, we performed secondary cleaning and manual verification to ensure high data quality while strictly adhering to MM2W's original train/test splits. Using this annotated data, we constructed a new evaluation set that treats generated low-level sub-tasks as direct input to evaluate the grounding performance of subsequent models, directly validating the effectiveness of DOM Tree Pruning Programming. Our final dataset contains approximately 5,000 high-quality interaction steps, divided into training and test sets.

\newcommand{\CommentLine}[1]{\STATE {\itshape // #1}}
\newcommand{\algcomment}[1]{\hfill \texttt{//} \textit{#1}}
\newcommand{\HYPER}{\item[\textbf{Hyperparameters:}]}

\begin{algorithm}[!t]
\caption{Scoring Function Template}
\label{alg:context_aware_scoring}
\begin{algorithmic}[1]

\renewcommand{\algorithmicrequire}{\textbf{Input:}}
\renewcommand{\algorithmicensure}{\textbf{Output:}}

\REQUIRE $E$: Pre-processed list of candidate elements; 
         \\$W$: Keywords with base weights ($\{k: w_{base}\}$) generated by LLM
\ENSURE $S$: Final relevance scores for ranking;
        \\$P$: Scoring justifications (paths)
% \STATE
% \CommentLine{Parameters:}
% \STATE $\alpha_{1} > \alpha_{2} > \alpha_{3} > \alpha_{4}$ \algcomment{Match Quality: Exact $>$ Phrase $>$ Word $>$ Fuzzy}
% \STATE $\beta_{1} > \beta_{2} > \beta_{3}$ \algcomment{Attribute Priority: Visual Text $>$ Trusted Attr $>$ Other Attr}
\HYPER 
    $\alpha_{1} > \alpha_{2} > \alpha_{3} > \alpha_{4}$ \algcomment{Match Quality: Exact $>$ Phrase $>$ Word $>$ Fuzzy} \\
    \hspace{\algorithmicindent} $\beta_{1} > \beta_{2} > \beta_{3}$ \algcomment{Attribute Priority: Visual Text $>$ Trusted Attribute $>$ Other Attribute}
% \STATE
\FOR{each element $e$ in $E$}
    \STATE $S[e] \leftarrow 0$; $P[e] \leftarrow \emptyset$
    % \CommentLine{Iterate over all content sources (text, attributes, tag):}
    \FOR{each attribute pair $(text, type)$ in $e$}
        
        % \CommentLine{Step 1: Determine Context Weight ($\beta$)}
        \IF{$type$ is Visual Text (e.g., $\text{text\_content}$)}
            \STATE $\beta \leftarrow \beta_{1}$
        \ELSIF{$type$ is Trusted Attribute (e.g., $\text{name}$)}
            \STATE $\beta \leftarrow \beta_{2}$
        \ELSIF{$type$ is Other Attribute (e.g., $\text{class}$, $\text{id}$)}
            \STATE $\beta \leftarrow \beta_{3}$
        \ENDIF
        
        % \CommentLine{Step 2: Match Keywords ($\alpha$) and Accumulate Score}
        \STATE $tokens \leftarrow \text{Split}(text)$
        
        \FOR{each keyword $k$ in $W$}
            \STATE $\alpha \leftarrow 0$
            
            % \CommentLine{Hierarchical Matching Logic}
            \IF{$text = k$}
                \STATE $\alpha \leftarrow \alpha_{1}$ \algcomment{full exact match}
            \ELSIF{($k$ has spaces) \textbf{and} ($k$ is substring of $text$)}
                \STATE $\alpha \leftarrow \alpha_{2}$ \algcomment{phrase match (on raw text)}
            \ELSIF{(\textbf{not} $k$ has spaces) \textbf{and} ($k \in tokens$)}
                \STATE $\alpha \leftarrow \alpha_{3}$ \algcomment{word match (on token list)}
            \ELSIF{$\text{FuzzyScore}(k, text, tokens) > \theta$}
                \STATE $\alpha \leftarrow \alpha_{4} \times \text{FuzzyScore}(k, text, tokens)$ \algcomment{fuzzy match}
            \ENDIF
            
            \IF{$\alpha > 0$}
                % \CommentLine{Accumulate: Base $\times$ Match Quality $\times$ Context Priority}
                \STATE $S[e] \leftarrow S[e] + (W[k] \times \alpha \times \beta)$
                \STATE $P[e].add(\text{Path}(k, type, \alpha))$ 
            \ENDIF
        \ENDFOR
    \ENDFOR
\ENDFOR
\RETURN $(S, P)$

\end{algorithmic}
\end{algorithm}

\subsection{Training Strategy}
\label{subsec:training_strategy}

Our training strategy involves two core stages: Supervised Fine-Tuning (SFT) and Reinforcement Fine-Tuning (RFT), both conducted on the Qwen2.5VL-3B-Instruct model.

\noindent\textbf{Supervised Fine-Tuning (SFT).}
The goal of SFT is to teach the base model to perform three distinct roles: Planner, Programmatic Element Filter, and Action Grounder. We explored two SFT paradigms: Separate Models and Unified Model. In the Separate Models approach, we fine-tune three independent models, each specialized for one task. The Planner maps high-level tasks and screenshots to low-level sub-tasks. The Programmatic Element Filter maps low-level sub-tasks to scoring program parameters. The Action Grounder maps low-level sub-tasks and pruned lists to final actions. In the Unified Model approach, we designed an innovative two-turn dialogue template to optimize a single model. In the first turn, the model acts as both Planner and Programmatic Element Filter, generating the low-level sub-task and scoring parameters simultaneously. In the second turn, after receiving the pruned list from the executed program, the model acts as the Action Grounder to output the final action. Experiments show the unified model is better suited for highly-coupled web automation tasks.

\noindent\textbf{Reinforcement Fine-Tuning (RFT).}
As SFT is insufficient for teaching complex, long-horizon planning and task decomposition, we employ Group Relative Policy Optimization (GRPO)~\cite{shao2024deepseekmath} for targeted RFT of the Planner (or the first turn of the unified model). We apply RFT selectively to the Planner because the Programmatic Element Filter and Action Grounder handle more deterministic tasks that can be effectively learned through SFT. The success of RFT depends on our Hierarchical Reward Mechanism, which provides timely feedback to the Planner based on downstream model performance. The Planner's reward $R_{total}$ at each step combines format and accuracy components: $R_{total} = R_{format} + R_{filtering} + R_{grounding}$. Here, $R_{format}$ ensures the generated low-level sub-task follows the correct format, $R_{filtering}$ provides critical intermediate feedback by verifying whether the Programmatic Element Filter's program successfully retains the ground-truth element in the pruned list, and $R_{grounding}$ measures final sub-task success based on the Action Grounder's output. In our design, these rewards are binary rewards (1 for success, 0 for failure).

\section{Experiments}
\label{sec:experiments}

\subsection{Experimental Setup}

\noindent\textbf{Benchmarks, Datasets, and Metrics.}
We conduct our primary offline evaluation on the standard Multimodal-Mind2Web benchmark~\cite{deng2023mind2web}, following its official evaluation metrics (Element Accuracy, Operation F1, and Step Success Rate). For model fine-tuning, we use a custom dataset of approximately 5,000 interaction steps created by re-annotating and cleaning the Multimodal-Mind2Web training and development sets (detailed in Section~\ref{subsec:main_results}). To assess the effectiveness of DOM Tree Pruning Programming, we build a new evaluation set from our re-annotated data. This benchmark uses ground-truth low-level sub-tasks as direct input to evaluate the grounding performance of the Programmatic Element Filter and Action Grounder models. We measure low-level sub-task grounding results using grounding accuracy. Additionally, we conduct targeted ablation studies on a curated set of online, dynamic websites, using LLM-Verified Task Completion Rate as the primary metric  in Section~\ref{subsec:ablation_studies}.

\noindent\textbf{Implementation Details.}
Our evaluation focuses on two versions of Prune4Web: a Two-turn Dialogue Unified version and a Separate Models version, both fine-tuned from Qwen2.5VL-3B-Instruct~\cite{bai2023qwen}. To assess low-level sub-task grounding performance, we also trained a lighter Qwen2.5-0.5B-Instruct~\cite{bai2023qwen} model, demonstrating that our Programmatic Element Filter and Action Grounder operate effectively with lightweight LLMs. We developed all Prune4Web models using the two-stage SFT+RFT training approach described in Section~\ref{subsec:ablation_studies}.

\noindent\textbf{Baselines.} We compare our method with proprietary models such as GPT-4~\cite{achiam2023gpt}, GPT-4o~\cite{hurst2024gpt}, and SeeAct~\cite{zheng2024gpt}, as well as state-of-the-art fine-tuning methods based on open-source models, including SeeClick-9.6B~\cite{cheng2024seeclick}, MiniCPM-3.1B~\cite{hu2024minicpm}, ScribeAgent-32B~\cite{shen2024scribeagent}, GPT-4o UGround~\cite{gou2024uground}, EDGE-9.6B~\cite{chen2024edge}, and MindAct Flan-T5XL~\cite{mind2web}.
% as well as a version utilizing GPT-4o as its backbone. Baselines for comparison include standalone GPT series models and other state-of-the-art models from the literature.

% \noindent\textbf{Training Paradigm.}
% All our fine-tuned Prune4Web models are developed using the two-stage SFT+RFT training paradigm detailed in Section 4.4.

\subsection{Main Results}
\label{subsec:main_results}

\noindent\textbf{Performance on Standard Web Benchmarks.}
On the official Multimodal-Mind2Web test splits (results in Table~\ref{tab:understanding}), our proposed Prune4Web, particularly the Two-turn Dialogue unified model, demonstrates strong performance and significantly outperforms several baselines. Notably, our model achieves this competitive performance on a moderately sized training set of only $\sim$5,000 trajectories while directly processing raw, complex HTML. This demonstrates our method's excellent data efficiency and significant potential for improvement.

\begin{table*}[t]
\centering
% 1. 必须改为 \small (9pt)。AAAI 禁止使用 \footnotesize (8pt) 或更小用于表格文本 。
\small 

% 2. 官方建议使用 1mm 来压缩列间距 。
\setlength{\tabcolsep}{1mm}

% 3. 移除竖线以符合学术出版的标准风格（搭配 booktabs），同时避免 overfull boxes。
\begin{tabularx}{\linewidth}{Xccccccccc}
\toprule
\multirow{2}{*}{Method} & \multicolumn{3}{c}{Cross-Task} & \multicolumn{3}{c}{Cross-Website} & \multicolumn{3}{c}{Cross-Domain}\\
\cmidrule(lr){2-4} \cmidrule(lr){5-7} \cmidrule(lr){8-10}
 & Ele. Acc & Op. F1 & Step SR & Ele. Acc & Op. F1 & Step SR & Ele. Acc & Op. F1 & Step SR\\
\midrule
% 4. 移除了 \cellcolor，因为 AAAI 禁止在表格中使用颜色 [cite: 404]。
\multicolumn{10}{c}{\textit{Proprietary Models}} \\
\midrule
% GPT-3.5 & 19.4 & 59.2 & 16.8 & 14.9 & 56.5 & 14.1 & 25.2 & 57.9 & 24.1 \\
GPT-4~\cite{achiam2023gpt} & 40.8 & 63.1 & 32.3 & 30.2 & 61.0 & 27.0 & 35.4 & 61.9 & 29.7  \\
GPT-4o~\cite{hurst2024gpt} & 5.7 & 77.2 & 4.3 & 5.7 & 79.0 & 3.9 & 5.5 & \textbf{86.4} & 4.5 \\
SeeAct \cite{zheng2024gpt} &46.4&73.4&40.2 &38.0&67.8&32.4 &42.4&69.3&36.8\\
\midrule
\multicolumn{10}{c}{\textit{Open-Source Models}} \\
\midrule
SeeClick-9.6B~\cite{cheng2024seeclick} & 26.3 & 86.2 & 23.7 &21.9 & \textbf{82.9} & 18.8 & 22.1 & 84.1 & 20.2 \\
MiniCPM-3.1B~\cite{hu2024minicpm} &23.8 &\textbf{86.8} &20.8 &20.3 &81.7 &17.3 &17.9 &74.5 &14.6\\
ScribeAgent-32B \cite{shen2024scribeagent}  & 38.0 & 52.9 & 35.6 & 34.1 & 52.7 & 32.5 & 39.4 & 54.7 & 37.3\\
GPT-4o UGround~\cite{gou2024uground} & 47.7&--&--&46.0&--&--&46.6&--&--\\
EDGE-9.6B \cite{chen2024edge} & -- & -- & 30.0 &-- & -- & 21.1  & -- & -- & 22.4\\
MindAct Flan-T5XL~\cite{mind2web} & 55.1 & 75.7 & 52.0 & 42.0 & 65.2 & 38.9 & 42.1 & 66.5 & 39.6\\
\midrule
\multicolumn{10}{c}{\textit{Prune4Web Variants}} \\
\midrule
Prune4Web-3B (Separate Models)   & 46.0 & 83.4 & 42.2 & 43.0 & 77.3 & 37.8 & 42.2 & 84.4 & 40.6 \\
Prune4Web-3B (Two-turn Dialogue Unified)    & \textbf{58.4} & 84.1 & \textbf{52.4} & \textbf{50.2} & 81.2 & \textbf{44.9} & \textbf{49.2} & 84.4 & \textbf{46.1} \\
\bottomrule
\end{tabularx}

% 标题必须保持在表格下方 。字体大小由 sty 文件自动处理。
\caption{Performance on the \textit{Multimodal-Mind2Web} benchmarks across different methods. (Ele. Acc: Element Accuracy; Op. F1: Operation F1; Step SR: Step Success Rate). Two variants of our Prune4Web are evaluated. The Separate Models approach uses three independent models for the planner, filter, and grounder. The Unified Model approach uses an innovative two-turn dialogue training strategy to optimize three stages as a single model. Top-1 accuracy is represented using \textbf{bold text}.}
\label{tab:understanding}
\end{table*}

\noindent\textbf{Performance on Low-Level Sub-Task Grounding.}
To precisely and isolatingly evaluate the effectiveness of DOM Tree Pruning Programming, we use a ground-truth low-level sub-task as direct input to evaluate the grounding performance of the Programmatic Element Filter and Action Grounder models. Since the unified Two-turn Dialogue model cannot be easily dissected for this purpose, we evaluate the Programmatic Element Filter and Action Grounder models trained using the Separate Models strategy. We report results for: 1) fine-tuning the Qwen2.5VL-3B-Instruct model using original HTML without pruning, 2) oracle pruning (GT elements guaranteed in top candidates), 3) direct pruning and decision with LLMs, and 4) our Prune4Web pruning and decision. The results (Table~\ref{tab:ablation_gt_plan_performance_v2}) show that, given a perfect low-level sub-task, our full Programmatic Element Filter–Action Grounder pipeline achieves a grounding accuracy of 88.28\%. This performance far surpasses the baseline without pruning (46.8\%) and significantly outperforms using the more powerful GPT-4o as the Action Grounder (80.65\%). Additionally, even with the much lighter Qwen2.5-0.5B-Instruct, our method shows superior performance on both pruning results and grounding accuracy. This experiment demonstrates that our DOM Tree Pruning Programming method achieves state-of-the-art performance in precise element localization and operation.

\begin{table}[t]
\centering
% 1. 使用 small (9pt) 字号，符合 AAAI 对表格字号的允许范围 (10pt 或 9pt)
\small 

% 2. 设置 tabularx 宽度为 \linewidth (单栏宽度)
\begin{tabularx}{\linewidth}{Xcc}
\toprule
% 使用嵌套 tabular 来换行，替代 \makecell，减少对非必要宏包的依赖
Method & Recall@20 & \begin{tabular}[c]{@{}c@{}}Grounding \\ Accuracy (\%)\end{tabular} \\
\midrule
% 3. 移除 \cellcolor，保留 \textit 用于区分组别
\multicolumn{3}{c}{\textit{GT Task + Original HTML (No Pruning)}} \\
\midrule
Qwen2.5VL-3B-instruct (FT) & -- & 46.80 \\
\midrule
\multicolumn{3}{c}{\textit{Oracle Pruning (GT guaranteed in top 20 candidates)}} \\
\midrule
GPT-4o & -- & 82.83 \\
GPT-4o-mini & -- & 75.39 \\
Qwen2.5VL-3B-instruct (ZS) & -- & 11.99 \\
Qwen2.5VL-7B-instruct (ZS) & -- & 12.08 \\
Qwen2.5VL-3B-instruct (FT) & -- & \textbf{90.28} \\
\midrule
\multicolumn{3}{c}{\textit{GT Task + End-to-End LLM Pruning \& Decision}} \\
\midrule
GPT-4o & 85.56 & 70.84 \\
GPT-4o-mini & 89.19 & 67.57 \\
\midrule
\multicolumn{3}{c}{\textit{GT Task + Prune4Web's Programmatic Element Filter Pruning}} \\
\midrule
GPT-4o & 85.56 & 80.65 \\
GPT-4o-mini & 89.19 & 73.75 \\
% MindAct & 97.15 & - \\
Qwen2.5-0.5B-instruct (FT) & \textbf{97.64} & \textbf{88.28} \\
Qwen2.5VL-3B-instruct (FT) & 97.46 & \textbf{88.28} \\
\bottomrule
\end{tabularx}

% 4. Caption 必须位于表格下方，且字号为 10pt (sty文件自动处理)
\caption{Performance with Ground-truth (GT) low-level sub-tasks on our custom grounding benchmark (1101 trajectories), evaluating Programmatic Element Filter and Action Grounder capabilities under various conditions. Recall@20 indicates the percentage of times the ground-truth element is successfully included within the top 20 candidates after the filtering stage. ZS denotes zero-shot, and FT denotes fine-tuning. Top-1 accuracy is indicated by \textbf{bold text}.}
\label{tab:ablation_gt_plan_performance_v2}
\end{table}

\subsection{Ablation Studies and Further Analyses}
\label{subsec:ablation_studies}

To meticulously validate the contributions of our key design choices, we conduct ablation studies and further analyses. These experiments investigate the precision of our filtering mechanism, the effectiveness of programmatic filtering compared to simpler baselines, the contribution of our multi-stage architecture, and the efficacy of our training strategies. We also evaluate the framework's robustness in dynamic online environments to demonstrate its practical applicability.

\noindent\textbf{Performance in Dynamic Online Environments.}
The effectiveness of Prune4Web in dynamic online environments is demonstrated through our ablation studies in Table~\ref{tab:ablation_pruning} and Table~\ref{tab:ablation_multi_agent}. Component analysis on a curated set of 30 online tasks shows consistent performance improvements. Our programmatic filtering significantly enhances task completion rates for smaller models like GPT-4o-mini, while the complete three-stage architecture achieves the best overall results. These findings confirm the framework's generalization capability and practical applicability in real-world settings.

\noindent\textbf{Effectiveness of Programmatic DOM Filtering.}
We compared our programmatic filtering against a baseline where the LLM directly performs Top-N selection in Table~\ref{tab:ablation_pruning}. The results show that for the powerful GPT-4o, our method maintains a high level of performance. However, its true value is demonstrated on smaller models. For GPT-4o-mini, Prune4Web's filtering boosts the task completion rate by over 5 percentage points (from 26.3\% to 31.6\%). For our fine-tuned Qwen2.5VL-3B, the baseline fails completely (0.0\%), while our structured method achieves a functional score (5.2\%). This highlights that the programmatic approach is essential for enabling smaller or specialized models to handle complex filtering tasks. filtering tasks.

\begin{table}[t]
\centering
% 1. 使用 \small (9pt)，符合允许的字号范围
\small

% 2. 移除 \resizebox，改用 tabularx 自动适应宽度
% 将第一列设为 X (自动换行/左对齐)，后三列居中
\begin{tabularx}{\linewidth}{Xccc}
\toprule
Filtering Method & GPT-4o & GPT-4o-mini & \begin{tabular}[c]{@{}c@{}}Qwen2.5\\VL-3B\end{tabular} \\
\midrule
LLM Top-N Selection & 42.1 & 26.3 & 0.0 \\
Prune4Web Filtering & \textbf{42.1} & \textbf{31.6} & \textbf{5.2} \\
\bottomrule
\end{tabularx}

% 3. 修正 Caption 位置：必须位于表格下方 (AAAI 强制要求)
\caption{Ablation on Programmatic DOM Filtering (LLM-Verified Task Completion Rate \%).}
\label{tab:ablation_pruning}
\end{table}

\noindent\textbf{Contribution of the Multi-Stage Architecture.}
We evaluated the necessity of our three-stage Planning-Filtering-Action Grounding architecture. As shown in Table~\ref{tab:ablation_multi_agent}, each stage provides a clear benefit. For GPT-4o-mini, starting with only an Action Grounder yields 21.1\% task completion. Adding the Planner boosts this to 26.3\%, and further adding our Programmatic Element Filter brings the final performance to 31.6\%. This steady improvement validates each component's contribution and confirms the rationality of our complete multi-stage design.

\begin{table}[t]
\centering
% 1. 使用 \small (9pt)，符合 AAAI 允许的表格字号 (10pt 或 9pt)
\small

% 2. 移除 \resizebox，改用 tabularx 自动适应宽度
% X 列用于长文本自动换行，c 列用于数字居中
\begin{tabularx}{\linewidth}{Xcc}
\toprule
Architecture & GPT-4o & GPT-4o-mini \\
\midrule
Action Grounder Only & 36.8 & 21.1 \\
Planner + Action Grounder & 42.1 & 26.3 \\
Full Framework (Prune4Web) & \textbf{42.1} & \textbf{31.6} \\
\bottomrule
\end{tabularx}

% 3. 修正 Caption 位置：必须移动到表格下方
\caption{Ablation on multi-stage framework (LLM-Verified Task Completion Rate \%).}
\label{tab:ablation_multi_agent}
\end{table}

\noindent\textbf{Efficacy of Training Strategies.}
We also assessed the impact of the RFT in Table~\ref{tab:ablation_training_sft_rft}. The results show that adding RFT on top of SFT consistently and significantly improves the Planner's capabilities. For the Separate Models framework, RFT boosts the Step Success Rate (Step SR) from 37.9\% to 42.2\%. For the Two-turn Dialogue Unified model, RFT provides an even larger boost, from 46.5\% to 52.4\%. These results confirm that our synergistic RFT approach, which uses filtering success as a reward, effectively optimize the Planner's policy for both training paradigms.

\begin{table}[t]
\centering
% 1. 使用 \small (9pt)，符合 AAAI 允许的表格字号
\small

% 2. 移除 \resizebox，改用 tabularx。
% 布局策略：
% - 第一列 (l): 内容较短 ("SFT Only")，左对齐。
% - 第二列 (X): 内容最长 ("Two-turn Dialogue Unified")，设为 X 自动换行/填充宽度。
% - 第三列 (c): 数字居中。
\begin{tabularx}{\linewidth}{lXc}
\toprule
Training Strategy & Framework & Step SR (\%) \\
\midrule
SFT Only & Separate Models & 37.9 \\
SFT + RFT & Separate Models & \textbf{42.2} \\
\midrule
SFT Only & Two-turn Dialogue  & 46.5 \\
SFT + RFT & Two-turn Dialogue  & \textbf{52.4} \\
\bottomrule
\end{tabularx}

% 3. 修正 Caption 位置：必须位于表格下方
\caption{Ablation on training strategies (Step SR: Step Success Rate \%) with Qwen2.5VL-3B-instruct on offline Multimodal-Mind2Web Cross-Task subset.}
\label{tab:ablation_training_sft_rft}
\end{table}

\begin{figure}[!t]
    % \centering
    \captionsetup{type=figure} \includegraphics[width=\linewidth]{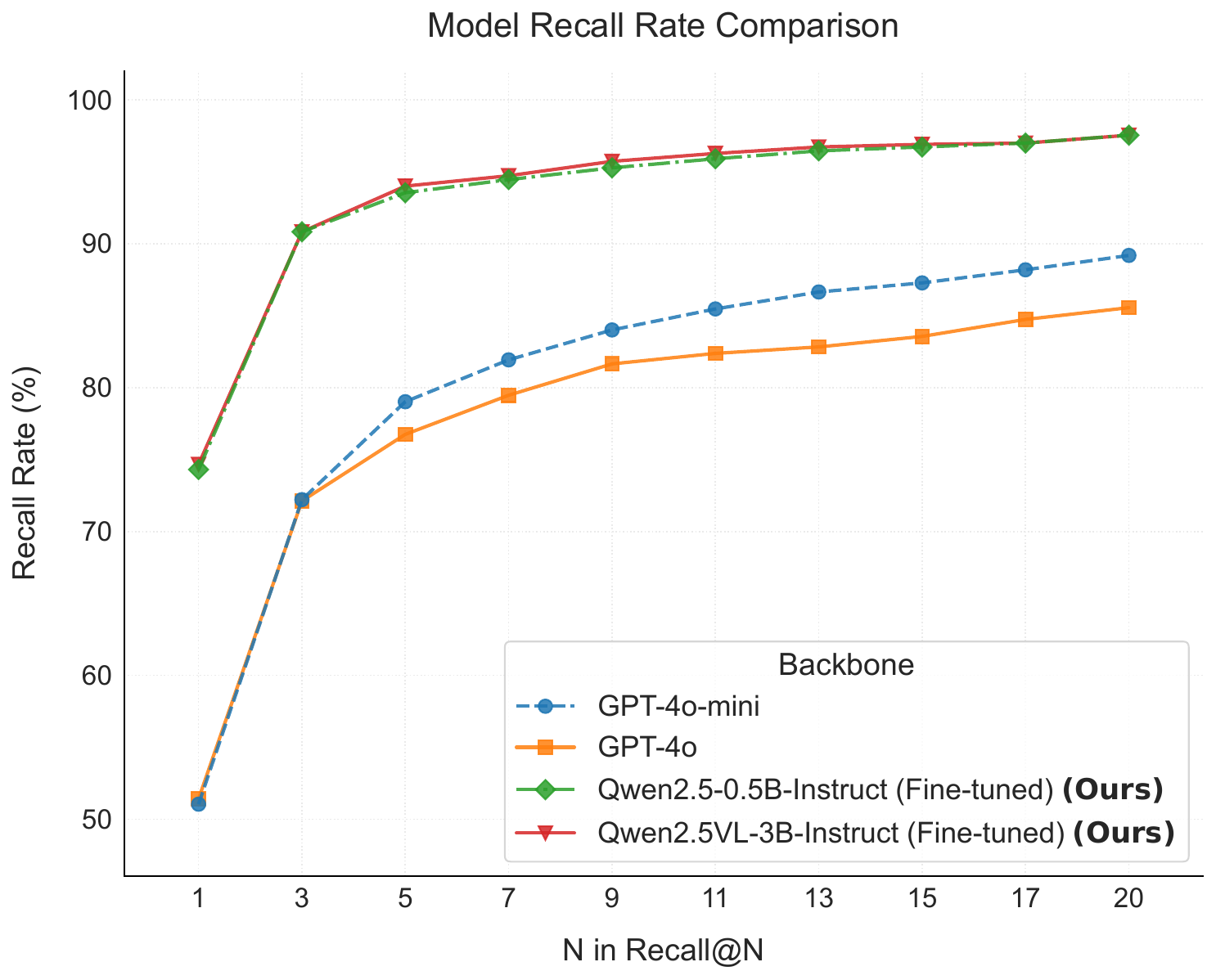}
    \captionof{figure}{Recall@N performance of our programmatic filtering stage with different backbone models. The y-axis represents the percentage of times the ground-truth element was successfully included in the Top-N candidates.}
    \label{fig:recall}
\end{figure}

\noindent\textbf{Filtering Recall Analysis.}
To evaluate the effectiveness of the scoring programs generated by the Programmatic Element Filter, we measured the Recall@N performance across various backbone models, as shown in Figure~\ref{fig:recall}. The results clearly indicate that our fine-tuned models significantly outperform the zero-shot GPT models at all values of $N$. Specifically, both our fine-tuned Qwen2.5-0.5B and 3B models achieve a recall rate of over 90\% when considering just the top 3 candidates ($N=3$), and approach 95\% at $N=5$. In contrast, the powerful GPT-4o model only reaches approximately 72\% recall at $N=3$ and ends at around 86\% at $N=20$. A particularly noteworthy finding is that our fine-tuned 0.5B model performs almost identically to the 3B model. This demonstrates that our DOM Tree Pruning Programming paradigm effectively distills the complex filtering task into a simple program generation problem that can be mastered even by smaller, more efficient models. This high recall with a small $N$ is crucial, as it provides the downstream Action Grounder with a small, high-quality set of candidates, significantly reducing the difficulty of the final grounding step.

\section{Conclusion}
\label{sec:conclusion}

This paper addressed the significant challenge of information overload for LLM-based web agents by introducing Prune4Web, a multi-stage architecture based on a Planning $\rightarrow$ Programmatic Filtering $\rightarrow$ Action Grounding workflow and the core method of \textbf{DOM Tree Pruning Programming}. Our key innovation leverages LLMs to generate lightweight, interpretable Python programs that dynamically score and prune DOM elements based on semantic clues from decomposed sub-tasks. This approach eliminates the need for LLMs to process massive DOMs, reducing candidate elements by 25$\sim $50 times while maintaining high accuracy. Our automated data annotation pipeline provides supporting data for training our model. Additionally, our two-turn dialogue training strategy jointly trains the Planner, Filter and Grounder as a unified model. This training approach combines SFT with a targeted RFT that uses intermediate filtering results as reward signals for the upstream Planner, significantly improving the model's strategic task decomposition capabilities. In conclusion, Prune4Web offers an effective and efficient solution to information overload through its innovative programmatic pruning paradigm and synergistic training strategy, laying a solid foundation for more accurate and reliable next-generation web automation systems.

\section{Acknowledgments}
This work was supported in part by National Natural Science Foundation of China (No.62461160331, No.62132001, No.62572039), in part by Huawei-BUAA Joint Lab, in part by the Fundamental Research Funds for the Central Universities, and in part by the Young Elite Scientists Sponsorship Program by CAST.

\bibliography{aaai2026}
\clearpage
\appendix
\input{sec/appendix}

\end{document}

%% file: sec/appendix.tex
\section*{Appendices}

This supplementary material provides additional details omitted in the main text to facilitate a deeper understanding of our methodology and for reproducibility. The appendices are organized as follows:

\begin{itemize}
\item \textbf{Appendix A: Framework Implementation Details.} This section elaborates on the technical implementation of the Prune4Web framework, including the internal data flow, the detailed logic of DOM Tree Pruning Programming, the complete prompts used for all models, and the definition of the agent's action space.

\item \textbf{Appendix B: Dataset Construction and Annotation.} This section provides a comprehensive overview of our data creation process. It details the selection and preprocessing of the source dataset, the annotation pipeline using GPT-4o, the multi-stage quality control measures, and the construction of our specialized benchmark for low-level sub-task grounding.

\item \textbf{Appendix C: Experimental Setup.} This section outlines all configurations required to reproduce our experiments. It covers the benchmarks and datasets used, the precise definitions of all evaluation metrics, the hardware and software environment, and the final hyperparameter settings for both SFT and RFT stages.

\item \textbf{Appendix D: Training Strategy Details.} This section offers an in-depth explanation of our training strategies. It describes the implementation of both the Separated Models and the Unified Model (Two-turn Dialogue) paradigms for SFT, and details the GRPO algorithm and the hierarchical reward mechanism used for RFT.

\item \textbf{Appendix E: Additional Experiments and Analyses.} This section presents supplementary experiments to further validate our method's efficiency and modularity. It includes a detailed analysis of filtering precision, a performance comparison using smaller-parameter models, and an experiment verifying the plug-and-play capability of our framework.

\item \textbf{Appendix F: Case Studies and Discussion.} This section provides an intuitive understanding of the framework's behavior through concrete examples. It includes a step-by-step walkthrough of a success case and an analysis of three distinct failure cases, followed by a systematic discussion of the method's limitations and directions for future work.

% \item \textbf{Appendix G: Reproducibility Checklist Clarifications.} This section provides a unified response to all items in the reproducibility checklist from the main paper and clarifies our plan for the public release of all associated code and datasets to ensure full reproducibility.

\end{itemize}

\section{Framework Implementation Details}
\label{sec:appendix_a}

This section provides the necessary technical details and specific parameters to supplement the main paper. Our goal is to ensure the complete reproducibility of the Prune4Web framework.

\subsection{Framework Overview}

As described in the main paper, the Prune4Web framework decomposes complex web tasks into a three-stage pipeline. The three stages are Planning, Filtering, and Action Grounding.

This section further clarifies the strong dependencies between these stages. The output of the Planning stage directly determines the goal for the Filtering stage. Correspondingly, the quality of the Filtering stage critically impacts the success of the Action Grounding stage.

This carefully designed structure of decoupled yet dependent components is key to the framework's ability to efficiently handle complex webpages. Its core philosophy is the separation of responsibilities. Each component focuses on the task it performs best. For example, the Planner is responsible for high-level strategic planning, while the Filtering and Grounding components handle low-level perception and execution.

To further clarify the system's interaction protocols and operational logic from an implementation perspective, we will now detail the internal data flow. Key data passed between models has explicit types. For instance, the low-level sub-task output by the Planner is a structured JSON object, while the parameters generated by the Programmatic Element Filter are passed to a Python function. The full DOM tree is provided as input only to the Filtering stage. In contrast, the Planner and Action Grounder process shorter, structured information. For example, the Planner receives a screenshot and a high-level task, while the Action Grounder receives a pruned list of candidate elements.

The workflow also contains explicit conditional logic. For example, the Filtering stage is activated only when the Planner's sub-task requires interaction with a page element. Otherwise, the workflow might proceed directly to an action that does not require element grounding, such as declaring the task complete in its final step. These implementation-level details are a necessary supplement to the framework diagram in the main text, ensuring that readers can understand the system's internal operating mechanics.

\subsection{DOM Tree Pruning Programming}

The main paper identifies DOM Tree Pruning Programming as our core technique to address the performance degradation of Large Language Models (LLMs) when processing long contexts. This section details its core implementation idea: we transform a complex, semantics-based filtering task into a simple code generation task based on a domain-specific language. Specifically, the LLM no longer parses the DOM directly. Instead, it generates a Python dictionary as a parameter, which is then passed to a fixed, lightweight scoring function that is subsequently executed in an external environment.

\noindent\textbf{Initial DOM Processing Script}.
Before executing the Python scoring function, we first run a JavaScript script, buildDomTree.js, in the browser environment to preprocess the live DOM. The core function of this script is to traverse the entire DOM tree and construct a simplified JSON object that contains only the elements potentially valuable for the current task. This JSON object then serves as the interactive\_elements input for the Python scoring function.

\noindent\textbf{Interactivity Judgment}.
The isInteractiveElement function within the script defines what constitutes an interactive element. This is a complex boolean logic that considers multiple dimensions. These dimensions include HTML tags such as \texttt{<a>} and \texttt{<button>}, ARIA roles like \texttt{button} and \texttt{link}, specific attributes such as \texttt{onclick} and \texttt{tabindex}, CSS styles like \texttt{cursor: pointer}, and event listeners detected via \texttt{window.getEventListeners}. The function also includes special handling for common web components like cookie consent banners to improve robustness in real-world web environments.

\noindent\textbf{Visibility and Hierarchy Judgment}.
In addition to interactivity, the script uses helper functions like isElementVisible and isTopElement to filter out elements that are invisible or obscured by other elements. This series of preprocessing steps ensures that the elements ultimately scored are those the user can currently see and interact with. It thus transforms a large, complex live DOM into a structured, pre-filtered, and smaller-scale JSON object, laying the foundation for efficient subsequent scoring.

\noindent\textbf{Scoring Function Template}.
The core of our method is a fixed, heuristic-based Python scoring function, the full implementation of which is provided in the supplementary code. Under this design, the LLM's task is significantly simplified. It only needs to generate a Python dictionary named keyword\_weights, where the keys are keyword strings and the values are their corresponding base weights, which are integers ranging from 1 to 50. The function relies on the rapidfuzz library for efficient fuzzy string matching and uses nltk.stem.PorterStemmer to normalize keywords and text through stemming.

\noindent\textbf{Complex Scoring Logic}.
The function employs a sophisticated, multi-layered weighted scoring mechanism to calculate the relevance score for each element. This includes base score bonuses for the highest-weighted keywords and for keywords matched in the text\_content, as well as match-type weighting for different match types like exact equality or phrase containment. It also uses the rapidfuzz library for fuzzy matching with defined thresholds and applies an attribute bonus for keywords matched in a predefined list of trusted attributes. This design hard-codes the complex and error-prone matching and scoring logic into the function template, ensuring the robustness and interpretability of the filtering process. At the same time, it constrains the LLM's task to a simple and well-defined generation problem.

\subsection{Model Prompt Engineering}

In the Prune4Web framework, the behavior of each LLM component is precisely guided by highly structured prompts. These prompts act as the bridge between our abstract methodology and the concrete behavior of the models. Their quality directly determines the framework's overall performance. To meet the highest standards of reproducibility, we provide the verbatim prompt templates used to generate the results reported in this paper in the following paragraphs.

\noindent\textbf{Planner Prompt}.
This prompt guides the model to act as a high-level planner. It requires the model to analyze the current situation based on the user's high-level task and a screenshot, and then generate a concise and clear low-level sub-task. The prompt includes strict output formatting requirements in JSON. It also contains specific instructions on how to handle common web interferences, such as pop-up dialogs.

\noindent\textbf{Programmatic Element Filter Prompt}.
This prompt configures the model as a keyword extraction expert. It receives the low-level sub-task from the Planner as input and is required to output a keyword\_weights dictionary containing keywords and their weights. The prompt provides strategic advice on weight assignment, such as assigning higher weights to key nouns, to guide the model toward generating high-quality scoring parameters.

\noindent\textbf{Action Grounder Prompt}.
This prompt requires the model to act as a precise executor. It receives the low-level sub-task and the pruned list of candidate elements as input. Its task is to select the single correct element from the list and decide on the action to be performed. The prompt emphasizes the need for logical reasoning during the thought process and to output the final decision in a specified format.

\noindent\textbf{``Two-turn Dialogue'' Unified Model Prompt}.
This prompt has a more complex structure and is used to train the unified model. In the first turn of the dialogue, the model acts as both the Planner and the Filter, generating the low-level sub-task and scoring parameters at once. After receiving the results from the externally executed filtering process, the dialogue proceeds to the second turn. In this second turn, the model then acts as the Action Grounder to make the final decision. This design aims to enhance the model's ability for continuous reasoning within the task's context.

\subsection{Action Space Definition}

To ensure the deterministic nature of the agent's decision outputs and the reliability of downstream execution, we define a discrete and strict action space. This action space can be considered the API protocol between the Action Grounder model and the browser interaction engine, which is based on Playwright. The actions, including their names, parameters, and functional descriptions, will be detailed in a table, as referenced in Table~\ref{tab:action_space}.

\noindent\textbf{Execution Details}.
The element\_index parameter is an abstract integer index. Upon receiving this index, the execution engine looks up the corresponding element's details, such as its XPath and CSS selectors, in the element hash map generated by buildDomTree.js. Subsequently, the engine employs a multi-strategy interaction method. It first attempts to perform a standard click or type operation using the retrieved selectors. If this fails, it falls back to dispatching JavaScript events, such as \texttt{element.dispatchEvent(new MouseEvent('click', ...))}, to simulate user interaction. This fallback mechanism is designed to improve the success rate of interactions on websites with complex front-end frameworks like React or Vue. Any model output that does not conform to the defined action space is considered an invalid action by the execution engine, thereby ensuring robust interaction.

\begin{table*}[t]
\centering
\caption{Definition of Action Space in the Prune4Web Framework}
\resizebox{\textwidth}{!}{%
\begin{tabular}{l|l|l}
\toprule
\textbf{Action} & \textbf{Description} & \textbf{Parameters} \\
\midrule
\texttt{CLICK(element\_id)} & Clicks on a specified element on the page. & \texttt{element\_id}: unique identifier from list $C_t$ \\
\texttt{TYPE(element\_id, text\_to\_input)} & Types the provided text into the specified input field. & \texttt{element\_id}, \texttt{text\_to\_input} \\
\texttt{SCROLL(direction)} & Scrolls the page vertically. & \texttt{direction}: ``up'' or ``down'' \\
\texttt{SELECT\_OPTION(element\_id, option\_value)} & Selects a specified option from a dropdown menu. & \texttt{element\_id}, \texttt{option\_value} \\
\texttt{NAVIGATE(url)} & Navigates to a new webpage. Used to start tasks or load new pages. & \texttt{url} \\
\texttt{DONE()} & Signals task/sub-task completion. & None \\
\texttt{FAIL()} & Indicates inability to complete the current task. & None \\
\bottomrule
\end{tabular}%
}

\label{tab:action_space}
\end{table*}

\section{Dataset Construction and Annotation}
\label{sec:appendix_b}

This section details the entire construction process for the custom dataset used to train and evaluate the Prune4Web framework. We aim to provide sufficient detail for other researchers to understand and reproduce our data preparation work.

\subsection{Data Source Selection and Preprocessing}

The effectiveness of the Prune4Web framework relies heavily on high-quality training data annotated with intermediate reasoning steps. Existing web automation datasets generally lack the fine-grained intermediate labels required by our three-stage model, necessitating that we construct our own.

\noindent\textbf{Primary Data Source}.
The foundation for all our training and evaluation data is the public Multimodal-Mind2Web~\cite{deng2023mind2web} (MM2W) dataset. Among the currently available datasets, MM2W is the only one that provides both rich HTML source code and corresponding webpage screenshots, making it suitable for our multimodal framework. Other datasets, such as WebVoyager~\cite{he2024webvoyager,he2024openwebvoyager} and Mind2Web-live~\cite{pan2024webcanvas}, only release their test sets and lack the necessary raw data for training. Therefore, MM2W is our sole choice at this stage, and we plan to collect or annotate more datasets in future work.

\noindent\textbf{Initial Data Cleaning}.
Before beginning the annotation process, we first conducted a comprehensive initial cleaning of the MM2W training and test sets to eliminate annotation errors present in the original data. The first cleaning criterion was to remove data points where the Ground-Truth (GT) Element was actually missing from the HTML source code. The second criterion involved processing the original HTML source with our buildDomTree.js script, as described in Appendix A. We then checked if the GT element existed in our constructed DOM tree. If the GT element was lost because it could not be assigned an ID or XPath, we also considered this an annotation error in the original dataset and removed the data point.

\noindent\textbf{Visual and Action Data Processing}.
The webpage screenshots provided in the MM2W dataset are often very long, with heights that can reach 10,000 pixels, making them unsuitable for direct model training and inference. We kept the screenshot width constant and vertically truncated the long screenshots into a series of standard-sized images with a height of 1080 pixels. We also recorded which of the truncated images contained the GT element. To enrich the action types, which were lacking in the original dataset, we added SCROLL and DONE actions. For truncated screenshots that did not contain the GT element, we annotated a SCROLL action. For the final screenshot in a task trajectory, we annotated a DONE action.

\subsection{Annotation and Quality Control with GPT-4o}

Our data annotation pipeline was designed to generate intermediate labels corresponding to the three core components of our framework for each interaction step in the MM2W dataset. To ensure absolute fairness in testing, we annotated \textbf{only the MM2W training set (cross\_train)}. The three official test sets (cross\_task, cross\_website, cross\_domain) were not additionally annotated and were used directly for the final end-to-end evaluation.

\noindent\textbf{Annotation Process}.
We used GPT-4o~\cite{hurst2024gpt} as the core annotation tool and designed specialized prompts for each annotation task. When annotating the low-level sub-task, we provided the GT element information to GPT-4o to ensure the high accuracy of the generated sub-task. In addition to the core labels, we also annotated extra attributes for each stage, such as the current\_state describing the page status and a detailed thinking process, to support potential future training based on richer contextual reasoning.

\noindent\textbf{Multi-Stage Quality Control}.
To ensure the final quality of the annotated data, we designed a rigorous multi-stage filtering process. The first stage was an automatic filter. After annotating the keywords and weights for the Programmatic Element Filter, we immediately invoked the scoring function to prune the DOM tree. If the GT element did not appear in the top-20 list of the pruned results, we considered the annotation or the original data to be of poor quality and completely removed that data point, including its corresponding Planner and Grounder annotations, from the training set. In the second stage of automatic filtering, during subsequent training preparation, we further removed data points with excessively long original HTML source tokens to avoid potential training issues. The third stage was manual verification. After all automatic filtering processes were complete, we conducted a final manual sampling check to ensure the overall high quality of the final dataset.

\subsection{Low-Level Sub-Task Grounding Benchmark}

To independently and precisely evaluate the effectiveness of our core method, DOM Tree Pruning Programming, we constructed a specialized benchmark from our newly annotated training set. We selected all steps that required interaction with page elements and randomly split them into a training set (80\%) and a test set (20\%). This benchmark is specifically designed to measure the performance of the subsequent Filtering and Grounding stages, given a perfect low-level sub-task.

\noindent\textbf{Evaluation Method}.
On this benchmark, we adopted a separate training strategy, independently retraining the Programmatic Element Filter and Action Grounder models. The evaluation process consists of two steps. In the first step, a model generates keywords and weights based on the input low-level sub-task and then performs pruning. In the second step, another model outputs the final action based on the pruned list of elements. This allows us to decouple the evaluation of the Planner's planning ability from the execution capability of the downstream components, thereby directly validating the effectiveness of our filtering and grounding methods.

\subsection{Dataset Statistics}

After all the cleaning, annotation, and filtering processes, we ultimately obtained a training dataset containing \textbf{5,503} high-quality interaction steps. An ``interaction step'' refers to a single user action within a multi-step task, such as a click or an input. Each step includes the complete annotation information required for the three stages of our framework.

\noindent\textbf{Dataset Splits and Usage}.
This dataset of 5,503 steps was further divided into two parts. A total of \textbf{4,402} steps (approximately 80\%) were used as the training set for the low-level sub-task grounding benchmark. The remaining \textbf{1,101} steps (approximately 20\%) were used as the test set for the same benchmark. The grounding accuracy of 88.28\% reported in the main paper was measured on these 1,101 test steps. It is important to note that the training set used for the end-to-end Prune4Web unified model consists of the entire dataset of 5,503 steps.

\section{Experimental Setup}
\label{sec:appendix_c}

This section provides the complete configuration details required to reproduce all experimental results in this paper, including the benchmarks used, precise definitions of evaluation metrics, the computing environment, and model training hyperparameters.

\subsection{Benchmarks and Datasets}

Our experimental design aims to comprehensively evaluate the performance of the Prune4Web framework across different scenarios. To this end, we employed both offline benchmarks and online dynamic websites for testing.

\noindent\textbf{Offline Evaluation on Multimodal-Mind2Web}.
Our main offline evaluation was conducted on the Multimodal-Mind2Web (MM2W) benchmark. We strictly adhere to its official splits for the training, validation, and three test sets (cross-task, cross-website, cross-domain). This ensures that our results can be fairly and directly compared with other work in the field.

\noindent\textbf{Online Evaluation on Dynamic Websites}.
To test the model's generalization ability and robustness in real, dynamic environments, we also conducted a series of online evaluations. We first filtered out a list of persistently inaccessible websites from the Mind2Web-live and WebVoyager test sets (e.g., kbb.com, sixflags.com, etc.). From the remaining pool of websites, we then randomly sampled 30 online tasks to form our test set, which covers common sites like Amazon and IMDb. We manually verified each of these 30 tasks to ensure they were completable during the testing period.

\noindent\textbf{Low-Level Sub-Task Grounding Benchmark}.
As described in Appendix B, we constructed a specialized benchmark to independently evaluate the effectiveness of DOM Tree Pruning Programming. This benchmark uses our annotated dataset and focuses on assessing the accuracy of the subsequent filtering and grounding stages, given a perfect low-level sub-task.

\subsection{Evaluation Metrics}

We used multiple metrics to evaluate model performance from different dimensions.

\noindent\textbf{Official MM2W Metrics}.
For the MM2W benchmark, we use its official metrics: Element Accuracy, Operation F1, and Step Success Rate (SR). These metrics respectively measure element localization; operation type and parameter filling; and the overall success of a single interaction step.

\noindent\textbf{Custom Grounding Metrics}.
For the low-level sub-task grounding benchmark, we use two main metrics. The first metric is Recall@k, which evaluates the performance of the filtering stage by measuring whether the ground-truth target element appears in the top-k candidates after pruning. The second metric is Grounding Accuracy, which evaluates the performance of the grounding stage by measuring whether the final executed action exactly matches the ground-truth action.

\noindent\textbf{LLM-Verified Task Completion Rate}.
For online evaluation, we use the LLM-Verified Task Completion Rate as the primary metric, as the final state of online tasks is difficult to judge with fixed rules. We employ GPT-4o as an automated evaluator. We provide GPT-4o with the original task instruction, along with the agent's sequence of observations (screenshots) and executed actions from the last N steps (e.g., N=3) of the trajectory. The prompt requires GPT-4o to determine if the core objective of the task has been achieved and to output a ``Success'' or ``Failure'' judgment with a brief rationale.

\subsection{Hardware and Software Environment}

\noindent\textbf{Hardware Infrastructure}.
All experiments were conducted on a server equipped with 8 NVIDIA A800 80G GPUs. The server's CPU is a 14-core processor with 100G of RAM. During training, each GPU process was bound to a dedicated CPU core to ensure efficiency.

\noindent\textbf{Software Stack}.
The software environment for the experiments was Ubuntu 20.04 with Python 3.10. Our training code was primarily developed based on the LLaMA-Factory~\cite{zheng2024llamafactory} framework.

\subsection{Hyperparameter Settings}

This section lists the final hyperparameter values for all models and algorithms in the SFT and RFT stages in a clear format. All reported hyperparameters were selected based on the best Step Success Rate achieved on the MM2W validation set.

\noindent\textbf{Supervised Fine-Tuning (SFT) Hyperparameters}.
We used Qwen2.5VL-3B-Instruct~\cite{bai2025qwen2.5} and Qwen2.5-0.5B-Instruct~\cite{bai2025qwen2.5} as our base models. The key hyperparameters for the SFT stage were as follows: we used the AdamW optimizer with a learning rate of 5.0e-5 and a cosine learning rate schedule. The batch size per device was 1, with 8 gradient accumulation steps, resulting in an effective global batch size of 64. We used bf16 mixed-precision for training and enabled flash\_attn for efficiency. The number of training epochs for all models was 3.

\noindent\textbf{Reinforcement Fine-Tuning (RFT) Hyperparameters}.
In the RFT stage, we used the GRPO~\cite{guo2025deepseek} algorithm to optimize the Planner (or the first turn of the unified model). We adopted the default hyperparameter settings from the VLM-R1~\cite{shen2025vlm} framework. In each optimization step, we generated K=4 responses for each prompt and scored them according to the reward function defined in Appendix D.

\section{Training Strategy Details}
\label{sec:appendix_d}

This section delves into the specific details of the Supervised Fine-Tuning (SFT) and Reinforcement Fine-Tuning (RFT) stages, which together form our complete training strategy.

\subsection{Supervised Fine-Tuning (SFT)}

All models' foundational capabilities are initialized through Supervised Fine-Tuning (SFT). The goal of SFT is to teach the model to generate outputs that conform to our predefined format and content, based on a given input. This stage provides a good ``cold start'' model for the subsequent Reinforcement Fine-Tuning (RFT) phase, allowing it to begin exploration from a reasonable base policy. We explored two SFT training paradigms: separated models and a unified model.

\noindent\textbf{Separated Models Paradigm}.
In the separated models paradigm, we independently trained three specialized models for the framework's three core components: the Planner, the Programmatic Element Filter, and the Action Grounder. Each model was trained on data specific to its corresponding task, and their respective training objectives and input-output formats are detailed below.

\noindent\textbf{Planner Model SFT}.
The training for the Planner model aims to teach it high-level planning. In training, the model receives input containing the current high-level task description and a webpage screenshot. It is trained to generate a structured string as output. This output string must contain two parts: a thought process enclosed in \texttt{<think>} tags, and a JSON object enclosed in \texttt{<answer>} tags. This JSON object needs to include a detailed analysis of the page state (\texttt{state\_analysis}), an evaluation of task progress (\texttt{progress\_evaluation}), challenges (\texttt{challenges}), the next low-level sub-task (\texttt{next\_steps}), the action type (\texttt{action\_type}), and the target text (\texttt{target}).

\noindent\textbf{Programmatic Element Filter Model SFT}.
The training for this model aims to teach it to extract keywords for element localization from the plan. During training, the model receives the complete JSON output from the Planner model as input. Its expected output is also a string containing \texttt{<think>} and \texttt{<answer>} tags. In this output, the JSON object enclosed by the \texttt{<answer>} tag must contain a \texttt{keyword\_weights} key, whose value is a dictionary mapping keywords to their weights.

\noindent\textbf{Action Grounder Model SFT}.
The training for the Action Grounder model aims to teach it to make the final decision based on the plan and the pruned list of elements. Its input is composed of the task context, the Planner's output, and the pruned list of DOM candidates. The model's expected output is also a string containing \texttt{<think>} and \texttt{<answer>} tags, where the JSON object enclosed by the \texttt{<answer>} tag must specify the final action type (\texttt{action}), the target element's ID (\texttt{id}), and any necessary input text (\texttt{input text}).

\noindent\textbf{Unified Model with Two-turn Dialogue}.
We designed the unified model paradigm to explore the trade-offs between two different optimization paths for Web Agent tasks: one path involves decomposing the task and optimizing multiple separated models, while the other utilizes a unified data stream to optimize a single, end-to-end model. This paradigm uses an innovative ``two-turn dialogue'' training template to synergistically optimize all three component functionalities within a single model. \textbf{First turn}: The model's input consists of the high-level task, history, and a webpage screenshot. Its expected output is a structured string containing a thought process in \texttt{<think>} tags, a low-level sub-task in \texttt{<plan>} tags, and a keyword-weight dictionary in \texttt{<keywords\_weights>} tags. \textbf{Second turn}: After the system simulates the filtering operation, the resulting pruned DOM list is provided as input. The model's expected output is then a JSON object containing the final action, enclosed in \texttt{<answer>} tags.

\noindent\textbf{Data Format Reference}.
The specific data formats and structures used in all the SFT training paradigms described above can be found in the data example files provided in the code repository accompanying our paper.

\subsection{Reinforcement Fine-Tuning (RFT)}

As SFT struggles to teach models complex, forward-looking, long-horizon planning, we employ Reinforcement Fine-Tuning (RFT) to further optimize the model. The goal of RFT is to enhance the Planner's strategic planning capabilities, enabling it to generate low-level sub-tasks that are more conducive to the success of downstream tasks. Therefore, RFT is applied only to the Planner model (in the separated paradigm) or to the first turn of the unified model.

\noindent\textbf{GRPO Algorithm Implementation}.
We use the GRPO (Group Relative Policy Optimization) algorithm for policy optimization. GRPO estimates the relative advantage of each response by comparing the rewards of multiple responses generated from a single prompt, which leads to more stable updates of the policy model. Our implementation is based on the VLM-R1 framework and uses its default hyperparameter configuration.

The GRPO algorithm extends policy gradient methods by refining the advantage estimation. A key aspect involves calculating the relative advantage $A_i$ for the $i$-th response in a set of candidate responses $O = \{o_1, o_2, \ldots, o_N\}$ generated from a state, given their respective rewards $\{r_1, r_2, \ldots, r_N\}$. The relative advantage is computed as:
$$A_i = \frac{r_i - \text{mean}(\{r_1, r_2, \ldots, r_N\})}{\text{std}(\{r_1, r_2, \ldots, r_N\}) + \epsilon_{std}}$$
where mean and std denote the mean and standard deviation of the rewards, and $\epsilon_{std}$ is a small constant for numerical stability. These relative advantages are then used to update the policy model, often under KL divergence constraints to stabilize training.

\noindent\textbf{Hierarchical Reward Mechanism}.
The key to successful RFT lies in the hierarchical reward mechanism we designed for the Planner. The core idea of this mechanism is that the quality of the Planner is determined not only by itself but, more importantly, by whether its generated plan enables the subsequent Filter and Grounder components to execute successfully. Thus, the reward for the Planner at step $t$, denoted as $R_t$, is composed of the format reward for its own output and the success rewards achieved by the downstream components in that step.

\noindent\textbf{Reward Function Definition}.
The reward we designed is step-wise; failure at one stage results in a reward of zero for all subsequent stages. The total reward $R_t$ is calculated with the formula: $R_t = R_{fmt} + \alpha \cdot R_{filtering\_success} + \beta \cdot R_{grounding\_success}$. Each component herein is a deterministic binary reward (1 for success, 0 for failure). \textbf{Format Reward ($R_{fmt}$)}: This checks if the Planner's output is a valid JSON object that contains all the required keys. The reward is 1 if the format is completely correct, and 0 otherwise. \textbf{Filtering Success Reward ($R_{filtering\_success}$)}: This reward is calculated only if $R_{fmt}=1$. The system executes the filtering operation and checks if the ground-truth target element appears in the top-20 list of pruned candidates. The reward is 1 if it is present, and 0 otherwise. \textbf{Grounding Success Reward ($R_{grounding\_success}$)}: This reward is calculated only if $R_{filtering\_success}=1$. The system proceeds with the grounding operation and checks if the final generated action perfectly matches the ground-truth action (including action type, element, and parameters). The reward is 1 if they match, and 0 otherwise. Considering this step-wise reward mechanism, we ultimately set the weight coefficients $\alpha$ and $\beta$ to 1 to ensure the model can optimize its performance stably and progressively.

\section{Additional Experiments and Analyses}
\label{sec:appendix_e}

This section provides additional experimental results and analyses to offer more detailed empirical evidence for the conclusions drawn in the main paper, and to further explore the efficiency and modularity of our method.

\subsection{Filtering Precision Analysis}

To more comprehensively demonstrate the precision of our DOM Tree Pruning Programming method, this section provides detailed performance data for the filtering stage, specifically the Recall@k metric. This data serves as a tabular supplement to Figure 3 in the main paper, allowing readers to look up the precise recall rate for each value of k.

\noindent\textbf{Detailed Recall@k Performance}.
Table~\ref{tab:recall_k_detail} details the Recall@k performance of four different backbone models: GPT-4o, GPT-4o-mini, Qwen2.5-0.5B-Instruct (Finetuned), and Qwen2.5VL-3B-Instruct (Finetuned), for k values ranging from 1 to 20. The data clearly shows that our fine-tuned models, even the smallest 0.5B version, exhibit extremely high recall rates at small k values (e.g., k <= 5), exceeding 95\%. This finding is particularly important as it demonstrates that our method can place the ground-truth target element at the very top of the candidate list with high precision, thereby significantly reducing the decision-making difficulty for the downstream Action Grounder model.

\begin{table}[h!]
\centering
\scriptsize
\caption{Sampled Recall@k performance of the filtering stage with different backbone models. Our fine-tuned models show superior performance, especially at smaller k values. All values are recall percentages (\%).}
\label{tab:recall_k_detail}
\begin{tabular}{lccccc}
\toprule
Model & 1 & 3 & 5 & 10 & 20 \\
\midrule
GPT-4o & 51.41 & 72.12 & 76.75 & 82.29 & 85.56 \\
GPT-4o-mini & 51.04 & 72.21 & 79.02 & 84.38 & 89.19 \\
MindAct~\cite{deng2023mind2web} & 51.10 & 79.74 & 87.48 & 94.20 & 97.15 \\  
Qwen2.5-0.5B (Ours) & 74.30 & 90.83 & 93.55 & 95.64 & 97.55 \\
Qwen2.5VL-3B (Ours) & \textbf{74.66} & \textbf{90.83} & \textbf{94.01} & \textbf{95.82} & \textbf{97.55} \\
\bottomrule
\end{tabular}
\end{table}

\subsection{Small-Parameter Model Performance Comparison}

To validate the efficiency of our method and its low dependency on computational resources, we designed an experiment to evaluate the performance when using a smaller-parameter model for the downstream components. Our hypothesis is that because the filtering and grounding tasks are greatly simplified within our framework, a smaller model should be able to maintain a high level of performance.

\noindent\textbf{Experimental Setup}.
We created a mixed-parameter model configuration. The Planner continued to use the Qwen2.5VL-3B-Instruct model to ensure planning quality. The downstream Programmatic Element Filter and Action Grounder, however, used our trained Qwen2.5-0.5B-Instruct model. We directly compare the performance of this mixed-model (3B Planner + 0.5B Downstream) with the separated model (Prune4Web-3B, Separated Models) reported in the main paper, which uses the 3B model for all components. The evaluation was conducted on the cross-task test set.

\noindent\textbf{Results and Analysis}.
Table~\ref{tab:small_model_comparison} shows the performance comparison of the two configurations on the three key metrics of the cross-task test set. The experimental results show that the configuration using the 0.5B model for downstream tasks performs very closely to the configuration using the 3B model for all components, with only a very slight drop in performance. This result strongly demonstrates that our method successfully transforms complex perception tasks into simpler generation tasks, making them manageable even for small-parameter models. This highlights the high efficiency and cost-effectiveness of our approach for practical deployment.

\begin{table}[h!]
\centering
\scriptsize
\caption{Performance comparison on the cross-task test set between the full 3B model configuration and a mixed-parameter configuration with a 0.5B downstream model.}
\label{tab:small_model_comparison}
\begin{tabular}{lccc}
\toprule
Method & Ele. Acc  & Op. F1 (\%) & Step SR (\%) \\
\midrule
Prune4Web-3B (Separated) & 46.0 & 83.4 & 42.2 \\
Prune4Web (3B Planner + 0.5B) & 44.6 & 82.4 & 41.3 \\
\bottomrule
\end{tabular}
\end{table}

\subsection{Plug-and-Play Capability Verification}

This experiment aims to demonstrate that our filtering and grounding modules are not exclusively tied to our own Planner. Instead, they can serve as a universal, plug-and-play component to enhance the grounding performance of any existing Web Agent. To this end, we chose the current state-of-the-art GUI Agent model, UI-tars~\cite{qin2025uitars}, as our integration target and comparison baseline, evaluating its performance on web tasks.

\noindent\textbf{Experimental Setup}.
We first evaluated the end-to-end performance of the UI-tars model on the cross-task test set as our baseline. Next, we constructed a hybrid system. We directly parsed the `thought` generated by UI-tars during its decision-making process and used it as the low-level sub-task input for our pre-trained Programmatic Element Filter and Action Grounder models. By comparing the performance of these two approaches, we can clearly see the performance gain brought by integrating our modules.

\noindent\textbf{Results and Analysis}.
Table~\ref{tab:plug_and_play} shows the performance comparison between the original UI-tars and the ``UI-tars + Prune4Web'' hybrid system on the cross-task test set. The results indicate that after integrating our filtering and grounding modules, the system's overall performance improved significantly across all three metrics. This is primarily because our programmatic pruning method effectively addresses the issue of inaccurate localization that the original model faced when dealing with complex DOMs. This experiment fully demonstrates that Prune4Web's filtering and grounding components possess excellent modularity and generalization capabilities, allowing them to serve as a universal enhancement technology to empower other Web Agent frameworks that can output a plan with a reasoning process.

\begin{table}[h!]
\centering
\footnotesize
\caption{Performance comparison on the cross-task test set between the original UI-tars-1.5-7B model and a hybrid system integrated with Prune4Web's downstream modules.}
\label{tab:plug_and_play}
\begin{tabular}{lcc}
% \hline
\toprule
Method & Op. F1 (\%) & Step SR (\%) \\
\midrule
UI-tars (End-to-End) & 84.7 & 53.6 \\
UI-tars + Prune4Web &  85.3 & 54.9 \\
\bottomrule
% \hline
\end{tabular}
\end{table}

\section{Case Studies and Discussion}
\label{sec:appendix_f}

This section provides detailed case studies to intuitively demonstrate the operational flow of the Prune4Web framework in practical tasks. Based on these cases, we then discuss the limitations of the current method and directions for future work.

\subsection{Success Case Analysis}

To provide readers with a more intuitive understanding of the effectiveness of the Prune4Web framework across different task types, we present four success cases. We will first provide a detailed breakdown of an e-commerce task to showcase the complete workflow of the framework, followed by a brief introduction to several other cases to highlight its adaptability in various scenarios.

\begin{figure*}[h!]
    \centering
    % Replace 'case_underarmour.pdf' with the actual path to your figure PDF
    \includegraphics[width=\linewidth]{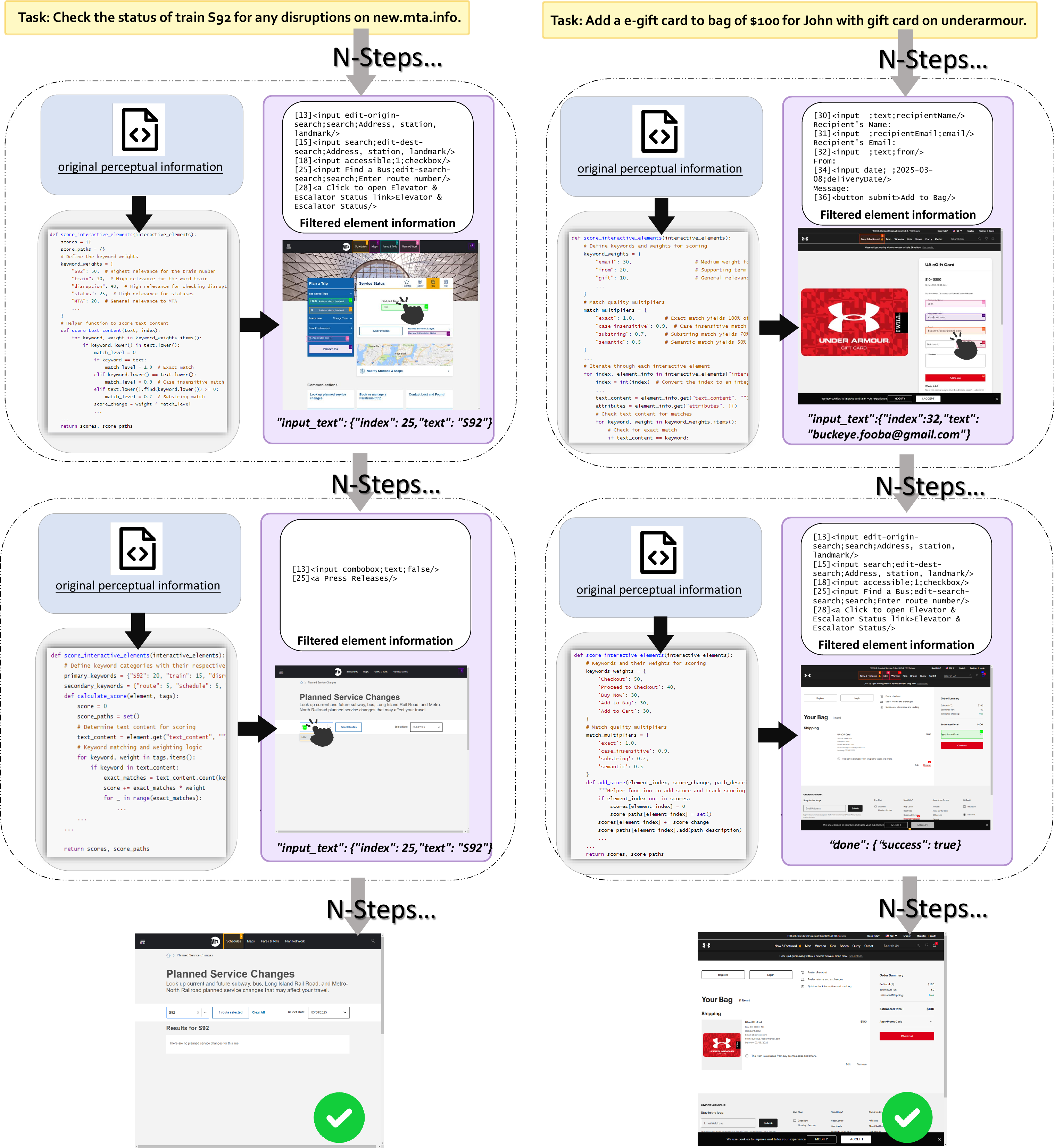}
    \caption{A detailed, step-by-step execution trace of the Prune4Web framework on an e-commerce gift card purchasing task. The figure highlights how programmatic pruning drastically reduces the candidate space at each step.}
    \label{fig:case_underarmour}
\end{figure*}

\noindent\textbf{Detailed Walkthrough: E-commerce Shopping Task}.
We use the task ``Buy a \$100 e-gift card for John on the Underarmour website'' as an example to demonstrate the end-to-end execution process of the framework. \textbf{Planning Stage}: After receiving the task and screenshot, the Planner accurately decomposes the task into a series of sub-tasks, such as ``fill in the recipient's email'' and ``click to add to bag.'' \textbf{Filtering Stage}: In each step, the Programmatic Element Filter generates highly relevant keywords (e.g., ``recipient,'' ``email'') based on the current sub-task. As shown in Figure~\ref{fig:case_underarmour}, the generated Python scoring program is executed, precisely pruning hundreds of interactive elements on the page down to just a few relevant input fields. \textbf{Grounding Stage}: The Action Grounder makes decisions on a very small set of candidates, easily selecting the correct element and performing the corresponding action, such as entering the email address. This case clearly demonstrates how our programmatic pruning method significantly reduces decision-making difficulty, thereby achieving precise and efficient multi-step form filling.

\begin{figure*}[h!]
    \centering
    % Replace 'case_amazon.pdf' with the actual path to your figure PDF
    \includegraphics[width=\linewidth]{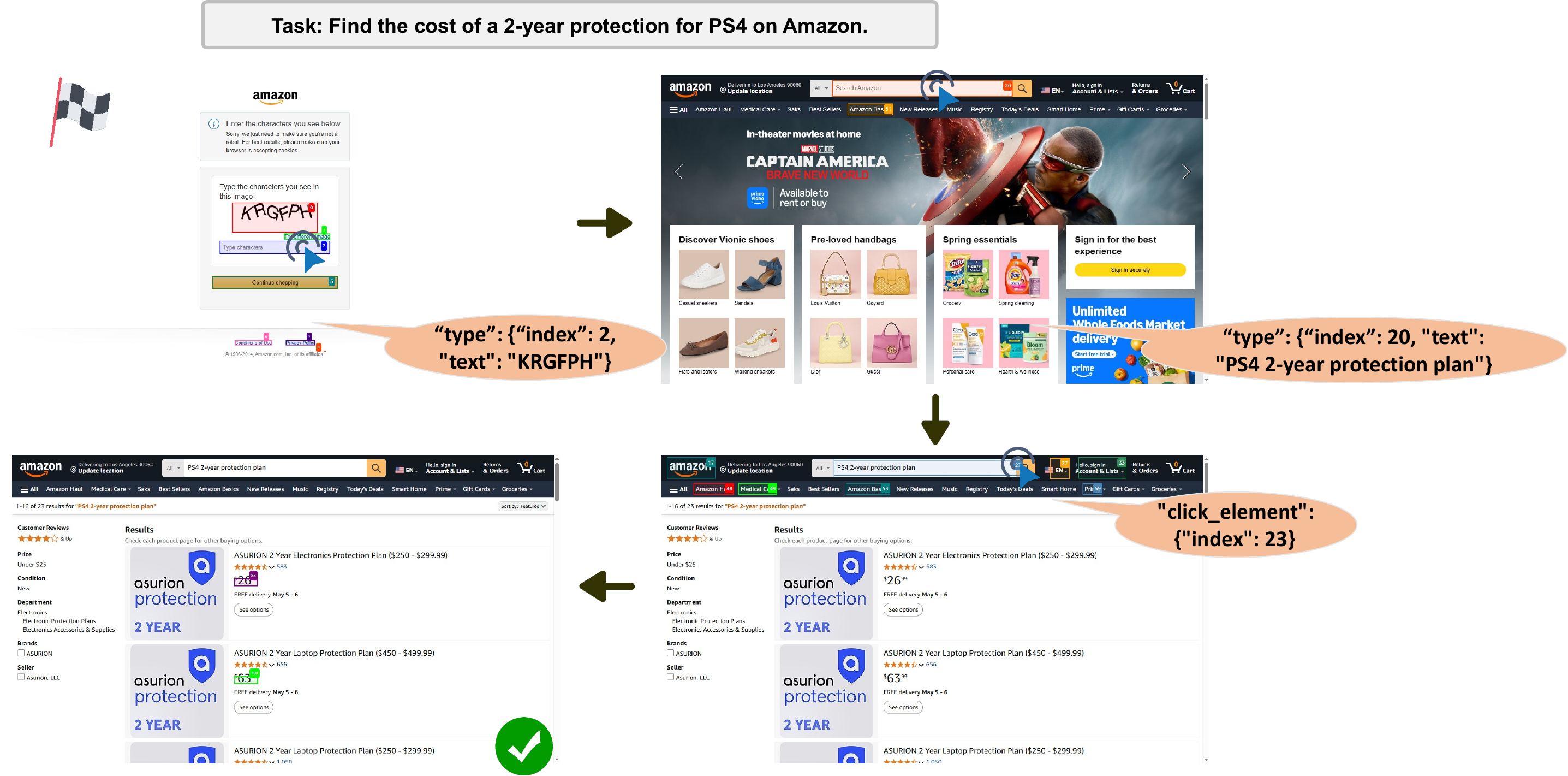}
    \caption{Partial task trajectory for finding a protection plan on Amazon. Our Programmatic DOM Pruning method filters a large number of irrelevant elements, leaving only a few highly relevant options and simplifying the agent's decision process.}
    \label{fig:case_amazon}
\end{figure*}

\noindent\textbf{Capability Showcase: Information Retrieval on Amazon}.
Prune4Web excels at information retrieval tasks that involve searching and navigating through results. In the task ``Find the cost of a 2-year protection for PS4 on Amazon,'' the agent must first handle a CAPTCHA, then search for the product, and finally identify the correct information from a list of results. As shown in Figure~\ref{fig:case_amazon}, the transition from a cluttered search results page to a focused view with only relevant protection plans clearly illustrates the effect of our Programmatic DOM Pruning. A large number of irrelevant elements are filtered out, simplifying the decision process and increasing the precision of the final action.

\begin{figure*}[h!]
    \centering
    % Replace 'case_github.pdf' with the actual path to your figure PDF
    \includegraphics[width=\linewidth]{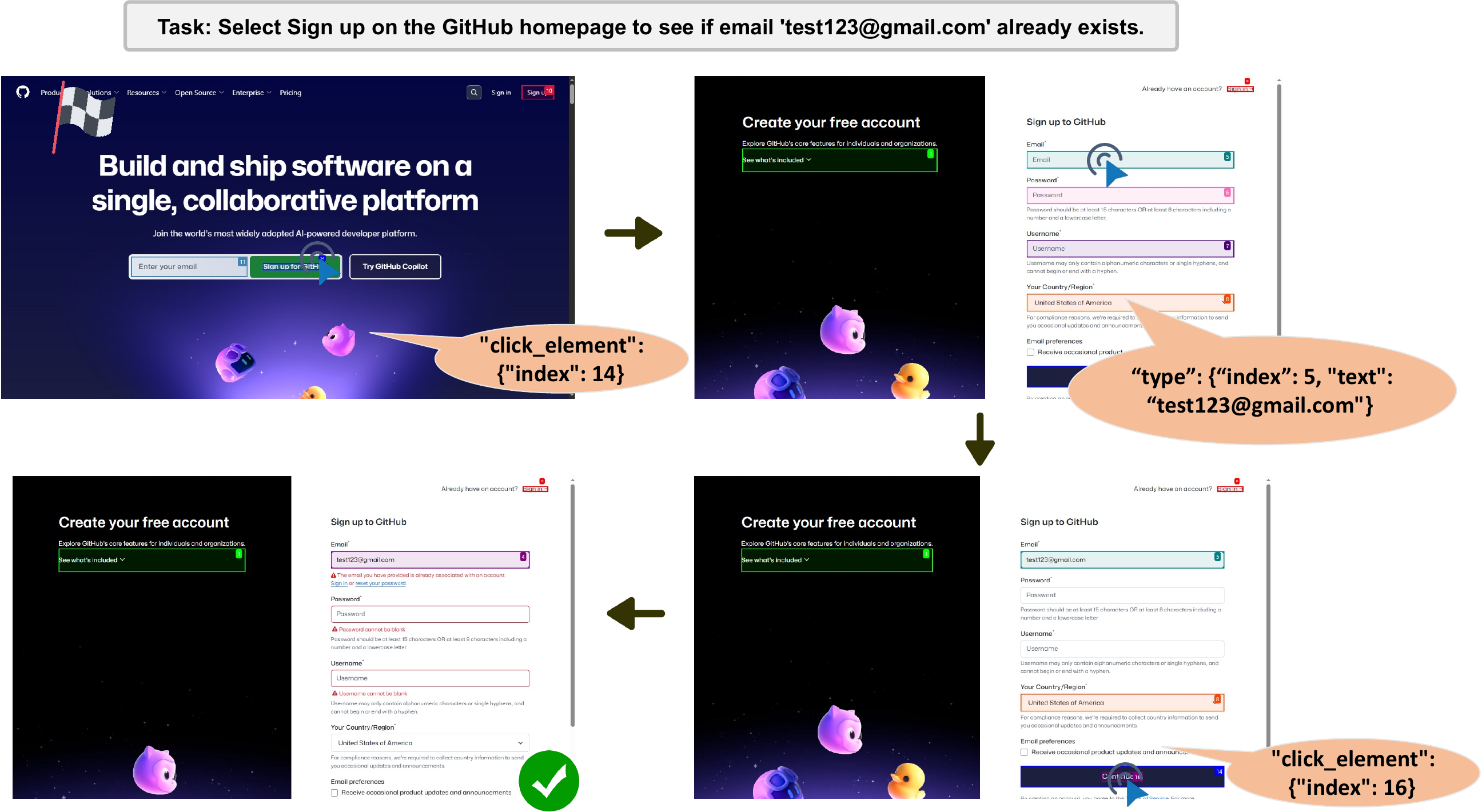}
    \caption{Dynamic adaptation of element pruning in the multi-step task of signing up for a GitHub account. Prune4Web's programmatic pruning dynamically adapts to the shifting focus of the agent at each step.}
    \label{fig:case_github}
\end{figure*}

\noindent\textbf{Capability Showcase: Handling Dynamic Webpages on GitHub}.
In tasks like ``Sign up for a GitHub account,'' the page state changes dynamically with user input. Our framework demonstrates good adaptability. As the low-level sub-task generated by the Planner shifts from clicking the initial ``Sign up'' button, to entering an email, and then to continuing, our programmatic pruning dynamically adapts to this shifting focus. As illustrated in Figure~\ref{fig:case_github}, at each distinct step, it precisely filters the webpage to highlight only the relevant interactive elements for the current sub-task, demonstrating the framework's adaptability in complex, multi-step scenarios.

\begin{figure*}[h!]
    \centering
    % Replace 'case_amtrak.pdf' with the actual path to your figure PDF
    \includegraphics[width=\linewidth]{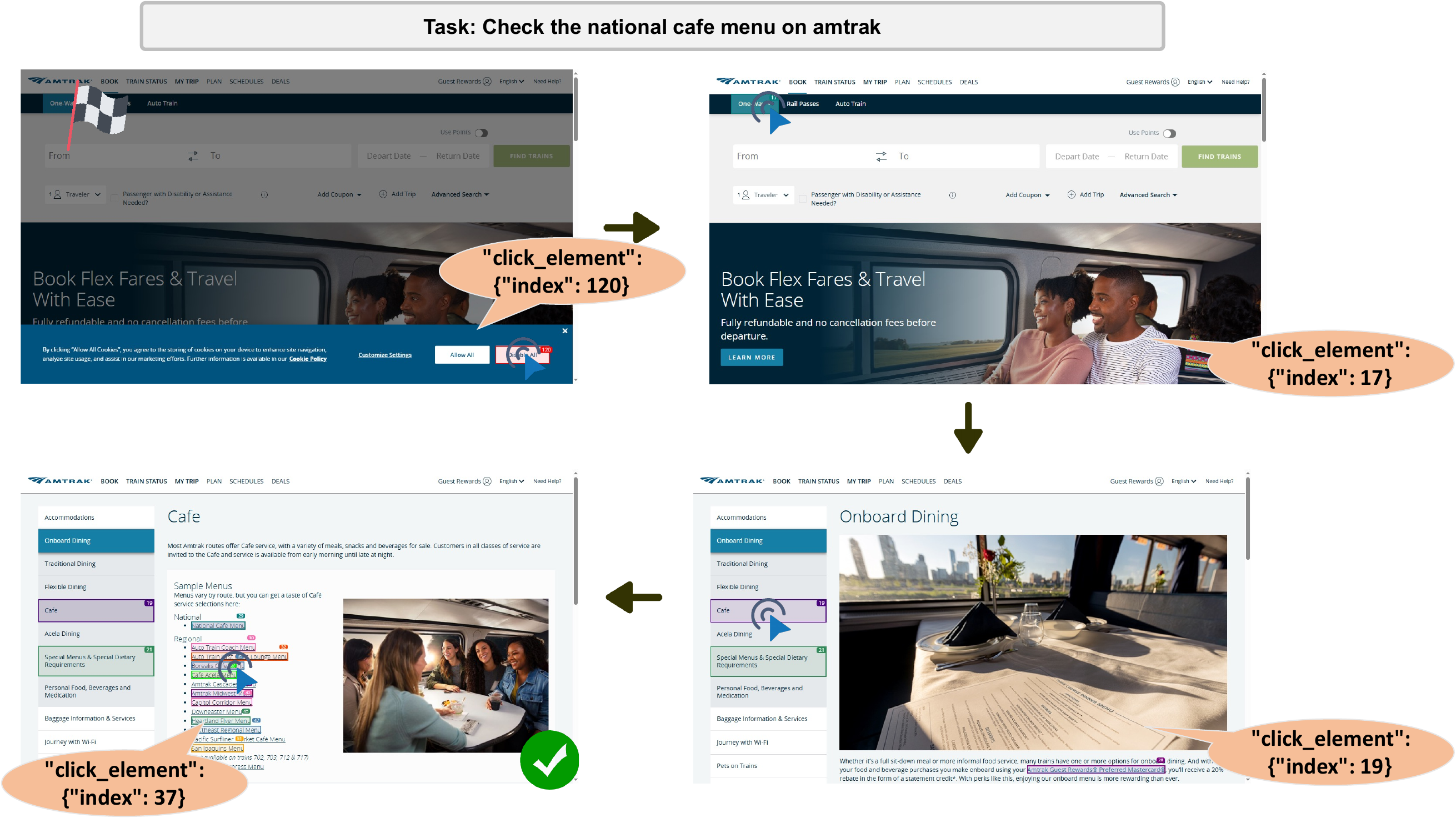}
    \caption{Navigational decision-making for an information retrieval task on Amtrak. The framework correctly identifies the semantically relevant hyperlink at each step to navigate to the correct page.}
    \label{fig:case_amtrak}
\end{figure*}

\noindent\textbf{Capability Showcase: Navigational Decision-Making on Amtrak}.
Unlike tasks that involve form-filling, many web automation challenges require the agent to make a series of navigational choices by understanding the semantic relevance of various hyperlinks. Figure~\ref{fig:case_amtrak} demonstrates the agent performing such a task: finding the national café menu on the Amtrak website. At each step, such as on the ``Onboard Dining'' page, the Prune4Web framework leverages the task's intent to programmatically prune the options, correctly identifying ``Café'' as the most relevant link to achieve the goal. This showcases the framework's effectiveness in handling complex navigational challenges by interpreting the semantic hierarchy of a website.

\subsection{Failure Case Analysis: The Interdependence of Planning and Grounding}

Analysis of failure cases reveals that the ultimate success of a task requires both high-quality Planning and precise Action Grounding. Although our Prune4Web framework performs exceptionally well in the Action Grounding stage, effectively mitigating the problem of inaccurate element localization, a severe deficiency in the upstream Planning stage can still lead to the ultimate failure of the task. We will elaborate on this point with two typical planning failure cases.

\noindent\textbf{Case 1: Ineffective Exploration Loop on Rottentomatoes}.
In the task ``View all of the Most Popular TV on rottentomatoes,'' the agent got stuck in a 25-step ineffective exploration loop and ultimately failed. The root cause of the failure lies in the Planner generating a flawed plan. It failed to identify the correct ``View All'' or ``Most Popular'' navigation links on the page, instead incorrectly guiding the agent to cycle through different category filters. This case demonstrates that a flawed plan, even if every step is executed with precision, cannot lead the agent to the final goal.

\begin{figure*}[h!]
    \centering
    % Replace 'case_rottentomatoes.pdf' with the actual path to your figure PDF
    \includegraphics[width=\linewidth]{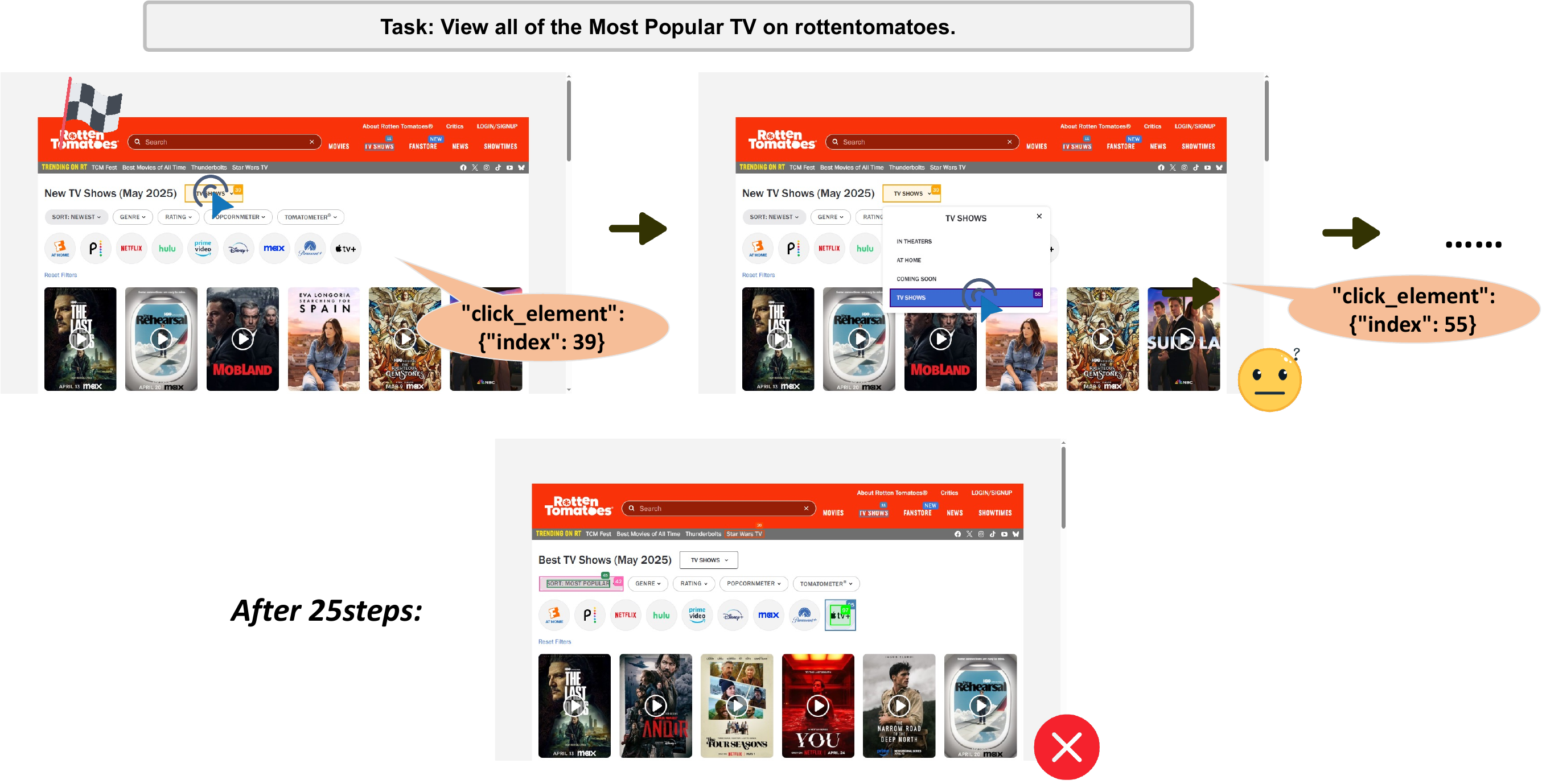}
    \caption{A failure case on Rottentomatoes caused by a flawed plan from the upstream Planner, leading to a prolonged and ineffective exploration loop. This highlights the primacy of planning for task success.}
    \label{fig:case_rottentomatoes}
\end{figure*}

\noindent\textbf{Case 2: Incorrect Goal Identification on Carmax}.
In the task ``Search for a full-time job in sales in Springfield on carmax,'' the agent ended up on an incorrect page for an inventory position in Raleigh. The failure again originated from the Planning stage. The Planner failed to correctly associate the core entities ``sales'' and ``Springfield'' with the corresponding input fields or filters on the page, instead formulating an incorrect search strategy. Although Prune4Web dynamically adjusted the pruning based on the incorrect intermediate intent at each subsequent step, the entire task was trapped in an irrecoverable failure loop from the very beginning due to the initial planning error.

\begin{figure*}[h!]
    \centering
    % Replace 'case_carmax.pdf' with the actual path to your figure PDF
    \includegraphics[width=\linewidth]{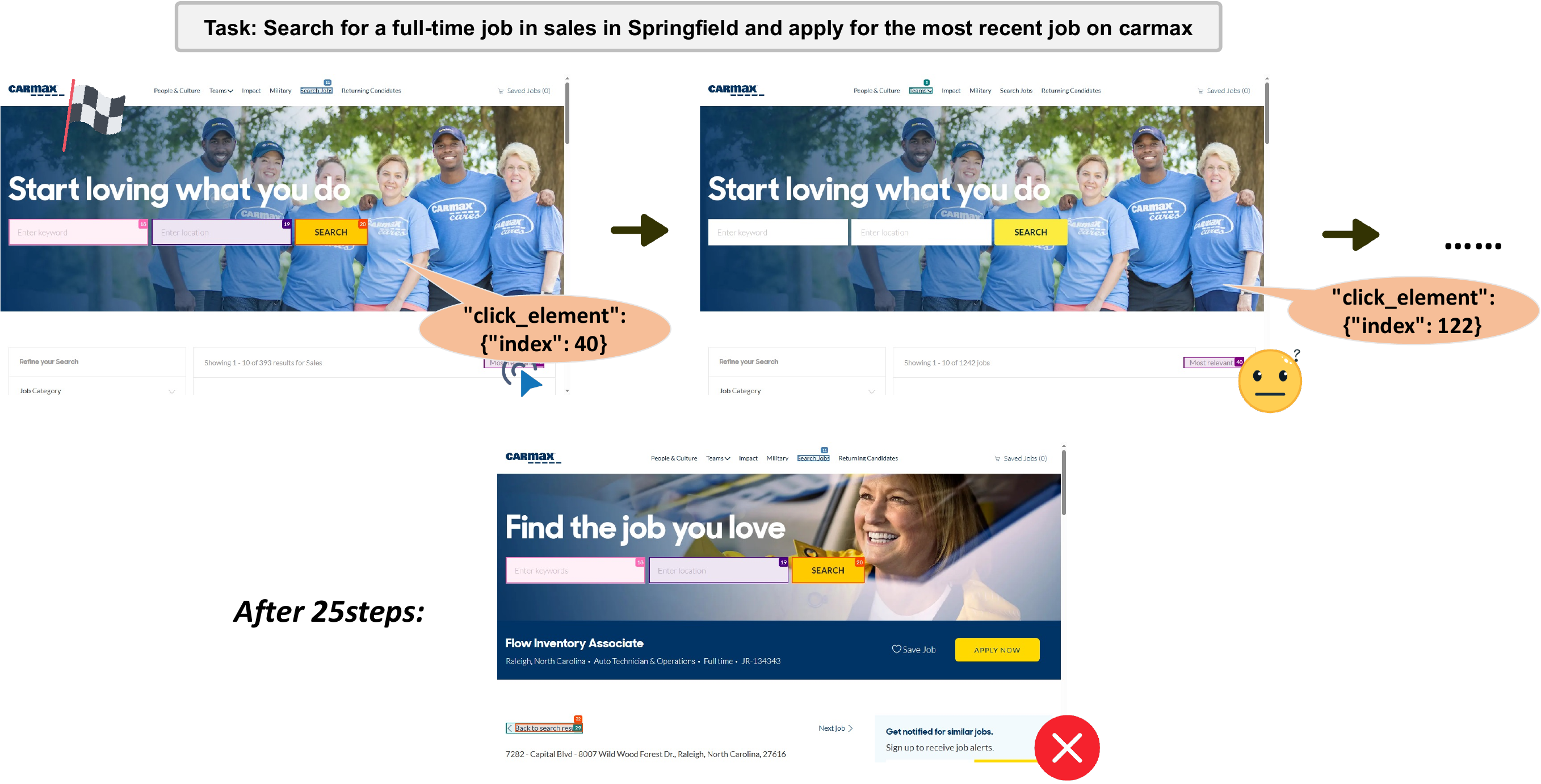}
    \caption{A failure case on Carmax where an initial planning error led the agent down a completely wrong path, which precise downstream execution could not rectify, reinforcing the critical role of upstream planning.}
    \label{fig:case_carmax}
\end{figure*}

\noindent\textbf{Common Conclusion}.
These two cases both point to a core conclusion: the precision of the downstream execution modules cannot compensate for the failures of the upstream planning. This once again confirms the critical importance of the planning module in the overall agent architecture.

\subsection{Limitations and Future Work}

Although Prune4Web achieved state-of-the-art performance in experiments, the aforementioned failure cases also reveal some limitations of its current implementation and help clarify the core contribution scope of this work. This section will systematically summarize these limitations and propose potential research directions accordingly.

\noindent\textbf{Primacy of Planning and Scope of This Work}.
As the failure cases show, the performance bottleneck of the current framework lies mainly in the Planning stage. The Planner may sometimes produce flawed or non-progressive plans, leading to task failure. Here, we wish to emphasize that the core contribution of the Prune4Web framework and method lies in the \textbf{optimization of the Action Grounding process}. We focus on solving the problem of how to accurately and efficiently localize and execute an action, given a low-level sub-task. The optimization of the planning and decomposition capabilities for high-level tasks, while crucial, is not the primary focus of this paper. Experiments show that for errors in the planning stage, optimization through SFT+RFT with a small amount of data is not significantly effective.

\noindent\textbf{Challenges in Filtering and Grounding}.
Although the current filtering and grounding stages are robust, they still face challenges when encountering non-standard web coding practices. \textbf{Non-standard HTML structure}: When web developers misuse non-semantic tags like \texttt{<div>} to build buttons or links, relying solely on CSS for their appearance and functionality, our method struggles to identify their interactivity due to the lack of explicit HTML tags or role attributes. \textbf{Lack of semantic features}: When a large number of interactive elements lack descriptive text, aria-labels, or other semantic attributes, our keyword matching mechanism becomes difficult. For example, a page might have many buttons with only icons and no text descriptions. \textbf{Visual and source code inconsistency}: Some websites prioritize visual presentation over the standardization of their HTML source code, leading to discrepancies between the text or structure in the source and what the user perceives visually. This can also mislead our filtering and grounding process.

\noindent\textbf{Future Work}.
Significantly improving planning capabilities can be an important direction for future work. This may require larger and more diverse planning training data, or the introduction of more powerful exploration mechanisms such as Monte Carlo Tree Search. To address the challenges in the filtering and grounding stages, future work could explore stronger multimodal fusion methods. For example, combining visual information to understand the functionality of elements defined by CSS, or using layout analysis to infer the actual function of non-semantic \texttt{<div>} tags, could help overcome the challenges posed by non-standard webpages. In summary, these limitations provide clear and valuable directions for future research in the field of web agents. We believe that a robust Web Agent requires the synergistic development of both planning and grounding capabilities.

% Note: Please ensure you have the 'tabularx' package included in your main LaTeX document preamble for these tables to render correctly.
% \usepackage{tabularx}

\begin{table*}[htbp]
\centering
\scriptsize
\caption{Planner Prompt Structure (Zero-shot Version)}
\label{tab:planner_prompt_zeroshot}
\begin{tabularx}{\textwidth}{@{}X@{}}
\toprule
\textbf{Planner System Prompt (Zero-shot)} \\
\midrule

\textbf{Role Definition:} \\
You are a planning agent that breaks down tasks into executable UI steps with strict safety protocols. Follow ABSOLUTELY:

\vspace{2mm}

\textbf{Core Rules:} \\
1. \textbf{POP-UP HANDLING:} Only close non-normal pop-ups that block or interfere with the task (e.g., ads, mandatory login walls, cookie consent dialogs). Do not close or dismiss any normal UI pop-ups that do not affect task execution (e.g., search suggestion dropdowns, informational tooltips).
2. \textbf{UI-ACTION FORMATTING:} Phrase steps as EXACT interface commands (e.g., ``click `Submit' button'', ``type `Paris' in search field'').
3. \textbf{LOGIN RESTRICTIONS:} NEVER trigger login UNLESS task explicitly mentions credentials or an undismissable login wall appears.
4. \textbf{TERMINATION CRITERIA:} TERMINATE SOLELY when: login wall appears WITH NO close option, paywall or other physical UI blockage occurs, or system security could be compromised.

\vspace{2mm}

\textbf{Available Actions:} \\
- Click elements \\
- Input text into forms \\
- Scroll page up/down \\
- Navigate to URLs (only for initial navigation)

\vspace{2mm}

\textbf{Output Format:} \\
\texttt{\{} \\
\quad\texttt{"state\_analysis": "Brief context analysis",} \\
\quad\texttt{"progress\_evaluation": "X\% - Description",} \\
\quad\texttt{"challenges": ["list"],} \\
\quad\texttt{"next\_steps": ["Only output ONE action. Format: click 'X' button | type 'Y' in Z field | TERMINATE if..."],} \\
\quad\texttt{"action\_type": "click | type | scroll | navigate",} \\
\quad\texttt{"target": \{} \\
\qquad\texttt{"text": "ELEMENT TEXT to interact with (e.g., 'Search box', 'Login button')"} \\
\quad\texttt{\},} \\
\quad\texttt{"reasoning": "Security/UI rationale"} \\
\texttt{\}}

\vspace{2mm}

\textbf{Critical Instructions:} \\
- For ALL actions: ``target.text'' must ONLY contain the TEXT OF THE ELEMENT to interact with \\
- For ``type'' actions: ``target.text'' is the element to type into (e.g., ``Search box''), NOT what you want to type \\
- The content to type should ONLY appear in ``next\_steps'', NOT in ``target.text''

\\
\bottomrule
\end{tabularx}
\end{table*}

\begin{table*}[htbp]
\centering
\scriptsize
\caption{Planner Prompt Structure (Training Version)}
\label{tab:planner_prompt_training}
\begin{tabularx}{\textwidth}{@{}X@{}}
\toprule
\textbf{Planner System Prompt (Training)} \\
\midrule

\textbf{Role Definition:} \\
You are a planning agent that breaks down tasks into executable UI steps. Output the thinking process in \texttt{<think>} ... \texttt{</think>} tags, and the final answer in \texttt{<answer>} ... \texttt{</answer>} tags.

\vspace{2mm}

\textbf{Output Format:} \\
\texttt{<think>} \\
\texttt{[Your thinking process here]} \\
\texttt{</think>} \\
\texttt{<answer>} \\
\texttt{\{} \\
\quad\texttt{"state\_analysis": "Brief context analysis",} \\
\quad\texttt{"progress\_evaluation": "X\% - Description",} \\
\quad\texttt{"challenges": ["list"],} \\
\quad\texttt{"next\_steps": ["Only output ONE action. Format: click 'X' button | type 'Y' in Z field"],} \\
\quad\texttt{"action\_type": "click | select | input | scroll",} \\
\quad\texttt{"target": \{} \\
\qquad\texttt{"text": "ELEMENT TEXT to interact with (e.g., 'Search box', 'Login button')"} \\
\quad\texttt{\}} \\
\texttt{\}} \\
\texttt{</answer>}

\\
\bottomrule
\end{tabularx}
\end{table*}

\begin{table*}[htbp]
\centering
\scriptsize
\caption{Programmatic Element Filter Prompt Structure (Zero-shot Version)}
\label{tab:filter_prompt_zeroshot}
\begin{tabularx}{\textwidth}{@{}X@{}}
\toprule
\textbf{Programmatic Element Filter System Prompt (Zero-shot)} \\
\midrule

\textbf{Role Definition:} \\
You are a professional filter keyword generator. Your task is to generate keywords with their corresponding weights for filtering and scoring interactive elements based on the complete plan output from the Planner. Output the thinking process in \texttt{<think>} ... \texttt{</think>} tags, and the final answer in \texttt{<answer>} ... \texttt{</answer>} tags.

\vspace{2mm}

\textbf{Task Description:} \\
Analyze the Planner's output and generate relevant keywords with appropriate weights that can be used to score and filter webpage elements for the given task.

\vspace{2mm}

\textbf{Keyword Weighting Strategy:} \\
- Assign higher weights (e.g., 30-50) to critical, task-specific keywords. \\
- Assign medium weights (e.g., 10-25) for supporting or contextual terms. \\
- Assign lower weights (e.g., 1-10) for general relevance terms.

\vspace{2mm}

\textbf{Output Format:} \\
\texttt{<think>} \\
\texttt{[Your keyword analysis and weight assignment thinking process]} \\
\texttt{</think>} \\
\texttt{<answer>} \\
\texttt{\{} \\
\quad\texttt{"keyword\_weights": \{} \\
\qquad\texttt{"keyword1": weight1,} \\
\qquad\texttt{"keyword2": weight2} \\
\quad\texttt{\}} \\
\texttt{\}} \\
\texttt{</answer>}

\\
\bottomrule
\end{tabularx}
\end{table*}

\begin{table*}[htbp]
\centering
\scriptsize
\caption{Programmatic Element Filter Prompt Structure (Training Version)}
\label{tab:filter_prompt_training}
\begin{tabularx}{\textwidth}{@{}X@{}}
\toprule
\textbf{Programmatic Element Filter System Prompt (Training)} \\
\midrule

\textbf{Role Definition:} \\
You are a professional filter keyword generator. Your task is to generate keywords and weights for scoring elements based on the provided plan.

\vspace{2mm}

\textbf{Task Description:} \\
Analyze the planner's output and generate relevant keywords with appropriate weights that can be used to score and filter webpage elements for the given task.

\vspace{2mm}

\textbf{Output Format:} \\
Directly output a JSON object with the following structure: \\
\texttt{\{} \\
\quad\texttt{"keyword\_weights": \{} \\
\qquad\texttt{"keyword1": weight1,} \\
\qquad\texttt{"keyword2": weight2} \\
\quad\texttt{\}} \\
\texttt{\}}

\\
\bottomrule
\end{tabularx}
\end{table*}

\begin{table*}[htbp]
\centering
\scriptsize
\caption{Action Grounder Prompt Structure (Zero-shot Version)}
\label{tab:grounder_prompt_zeroshot}
\begin{tabularx}{\textwidth}{@{}X@{}}
\toprule
\textbf{Action Grounder System Prompt (Zero-shot)} \\
\midrule

\textbf{Role Definition:} \\
You are an AI agent designed to automate browser tasks. Your goal is to accomplish the ultimate task by following the rules.

\vspace{2mm}

\textbf{Thinking Requirements:} \\
In your thinking process, carefully analyze both the Planner's results and the pruned DOM tree to identify the correct element. Specifically describe how you used the Planner's target information and the DOM tree to make your decision. \textbf{IMPORTANT}: Explicitly mention element IDs when discussing them (e.g., ``I identified element ID 123 as the target because...'' or ``Comparing elements with IDs 456 and 789, I determined that ID 456 is more appropriate because...'').

\vspace{2mm}

\textbf{Input Format:} \\
Elements are presented in a standard format with IDs: \texttt{[id]<type>text</type>} \\
- Only elements with numeric IDs in [] are interactive. \\
- Elements without [] provide only context.

\vspace{2mm}

\textbf{Output Format:} \\
\texttt{<think>} \\
\texttt{[Your thinking process with explicit element ID mentions]} \\
\texttt{</think>} \\
\texttt{<answer>} \\
\texttt{\{} \\
\quad\texttt{"action": "click\_element | input\_text | select\_element | scroll",} \\
\quad\texttt{"id": element\_id\_integer,} \\
\quad\texttt{"input text": "input\_text\_content or 'no input text' [default]"} \\
\texttt{\}} \\
\texttt{</answer>}

\\
\bottomrule
\end{tabularx}
\end{table*}

\begin{table*}[htbp]
\centering
\scriptsize
\caption{Action Grounder Prompt Structure (Training Version)}
\label{tab:grounder_prompt_training}
\begin{tabularx}{\textwidth}{@{}X@{}}
\toprule
\textbf{Action Grounder System Prompt (Training)} \\
\midrule

\textbf{Role Definition:} \\
You are an AI agent designed to automate browser tasks. Your goal is to accomplish the ultimate task by selecting the correct element from the provided list.

\vspace{2mm}

\textbf{Input Format:} \\
Elements are presented in a standard format with IDs: \texttt{[id]<type>text</type>} \\
- Only elements with numeric IDs in [] are interactive. \\
- Elements without [] provide only context.

\vspace{2mm}

\textbf{Output Format:} \\
Directly output a JSON object with the following structure: \\
\texttt{\{} \\
\quad\texttt{"action": "click\_element | input\_text | select\_element | scroll",} \\
\quad\texttt{"id": element\_id\_integer,} \\
\quad\texttt{"input text": "input\_text\_content or 'no input text' [default]"} \\
\texttt{\}}

\\
\bottomrule
\end{tabularx}
\end{table*}

\begin{table*}[htbp]
\centering
\scriptsize
\caption{LLM-Verified Task Completion Rate (LVCTR) Prompt}
\label{tab:lvctr_prompt}
\begin{tabularx}{\textwidth}{@{}X@{}}
\toprule
\textbf{Task Completion Verification Prompt} \\
\midrule

\textbf{Role Definition:} \\
You are an impartial evaluator. Your task is to determine if a web agent successfully completed its assigned task based on its final actions and observations.

\vspace{2mm}

\textbf{Task Instruction:} \\
\texttt{\{Original Task Description\}}

\vspace{2mm}

\textbf{Agent's Final Trajectory (Last 3 Steps):} \\
\texttt{Step N-2:} \\
\texttt{Observation: [Screenshot Image]} \\
\texttt{Action: \{action\_details\}} \\
\texttt{Step N-1:} \\
\texttt{Observation: [Screenshot Image]} \\
\texttt{Action: \{action\_details\}} \\
\texttt{Step N:} \\
\texttt{Observation: [Screenshot Image]} \\
\texttt{Action: \{action\_details\}}

\vspace{2mm}

\textbf{Your Task:} \\
Based on the final trajectory, did the agent successfully complete the core objective of the task instruction? Please answer with a JSON object containing two keys: ``success'' (boolean) and ``reasoning'' (a brief explanation for your decision).

\vspace{2mm}

\textbf{Output Format:} \\
\texttt{\{} \\
\quad\texttt{"success": true|false,} \\
\quad\texttt{"reasoning": "Your brief explanation here."} \\
\texttt{\}}

\\
\bottomrule
\end{tabularx}
\end{table*}

\begin{table*}[htbp]
\centering
\scriptsize
\caption{Two-turn Dialogue Training Template}
\label{tab:two_turn_dialogue_template}
\begin{tabularx}{\textwidth}{@{}X@{}}
\toprule
\textbf{Unified Model Training Conversation Structure} \\
\midrule

\textbf{System Message:} \\
You are an expert web automation assistant. Based on the user's task description, history, and a screenshot of the current webpage, you must first generate a thought process, a plan, and a set of keywords and weights for scoring DOM elements. Afterwards, when presented with a list of candidate elements, you must make the final selection.

\vspace{4mm}

\textbf{Turn 1} \\
\midrule
\textbf{User Input:} \\
\texttt{<image>\{"task\_description": "...", "history": [...] \}}

\vspace{2mm}

\textbf{Assistant Output (Expected):} \\
\texttt{<think>...</think> <plan>...</plan> <keywords\_weights>\{...\}</keywords\_weights>} \\
\midrule

\textbf{Turn 2} \\
\midrule
\textbf{User Input:} \\
\texttt{Here is the list of candidate elements, each with a unique ID. Please choose the ID of the correct element.} \\
\texttt{[id]<type>...</type>} \\
\texttt{[id]<type>...</type>} \\
\texttt{...}

\vspace{2mm}

\textbf{Assistant Output (Expected):} \\
\texttt{<answer>\{'action': ..., 'id': ..., 'input text': ...\}</answer>} \\
\bottomrule
\end{tabularx}
\end{table*}